\documentclass{article}
\usepackage{arxiv}

\usepackage{comment}
\usepackage{endnotes}

\usepackage{subfig}

\def\E{{\mathbb E}}

\def\d{\mathrm{d}}
\def\Tr{\mathrm{Tr}}

\usepackage{mathtools}
\newcommand{\defeq}{\vcentcolon=}
\newcommand{\eqdef}{=\vcentcolon}
\DeclareMathOperator*{\argmax}{argmax}
\DeclareMathOperator*{\argmin}{argmin}
\usepackage{booktabs}

\usepackage{graphicx}%
\usepackage{multirow}%
\usepackage{amsmath,amssymb,amsfonts}%
\usepackage{amsthm}%
\usepackage{mathrsfs}%
\usepackage[title]{appendix}%
\usepackage{xcolor}%
\usepackage{textcomp}%
\usepackage{manyfoot}%
\usepackage{booktabs}%
\usepackage{listings}%
\usepackage[ruled]{algorithm2e}
\usepackage{natbib}

\newtheorem{lemma}{Lemma} 
\newtheorem{assumption}{Assumption}

\usepackage{enumitem}

\newtheorem{theorem}{Theorem}
\newtheorem{remark}{Remark}%

\newtheorem{definition}{Definition}%

\title{Digital Twin Calibration with Model-Based Reinforcement Learning}

\author{Hua Zheng  \\
 Northeastern University
 \And
  Wei Xie \thanks{Corresponding author. Email: w.xie@northeastern.edu} \\
 Northeastern University
\And
  Ilya O. Ryzhov \thanks{Corresponding author. Email: iryzhov@umd.edu}\\
 University of Maryland
\And
  Keilung Choy\\
 Northeastern University
 }
\date{}
\begin{document}
\maketitle

\begin{abstract}
This paper presents a novel methodological framework, called the Actor-Simulator, that incorporates the calibration of digital twins into model-based reinforcement learning for more effective control of stochastic systems with complex nonlinear dynamics. Traditional model-based control often relies on restrictive structural assumptions (such as linear state transitions) and fails to account for parameter uncertainty in the model. These issues become particularly critical in industries such as biopharmaceutical manufacturing, where process dynamics are complex and not fully known, and only a limited amount of data is available. Our approach jointly calibrates the digital twin and searches for an optimal control policy, thus accounting for and reducing model error. We balance exploration and exploitation by using policy performance as a guide for data collection. This dual-component approach provably converges to the optimal policy, and outperforms existing methods in extensive numerical experiments based on the biopharmaceutical manufacturing domain.
\end{abstract}




\maketitle

\section{Introduction}
\label{sec:introduction}

{This study is motivated by optimal control applications that exhibit \textit{high complexity, high uncertainty, and very limited data} \citep{wang2024biomanufacturing,zheng2023policy,Plotkin2017,mirasol2017challenges}. In particular, all of these challenges are present in the domain of biopharmaceutical manufacturing, used for production of essential life-saving treatments for severe and chronic diseases, including cancers, autoimmune disorders, metabolic diseases, genetic disorders, and infectious diseases such as COVID-19 \citep{zahavi2020monoclonal, teo2022review}. Using cells as factories, biomanufacturing involves hundreds of biological, physical, and chemical 
factors dynamically interacting with each other at molecular, cellular, and macroscopic levels and impacting production outcomes. Due to the complexity of these mechanisms, it is quite difficult to control production safely and effectively, especially in the presence of very limited data.}



Digital twins have proven very useful in guiding the control of complex physical systems \citep{tao2018digital}. However, there are many situations where the system dynamics are not fully understood and only crudely approximated by digital twin models \citep{zheng2023policy,nguyen2011model}. For example, in biopharmaceutical manufacturing and cell culture processes, a digital twin, built only on limited knowledge of the critical metabolic mechanisms and kinetics of cells, can lead to suboptimal control strategies, such as oxygen and nutrient feeding, that eventually impact the production outputs (e.g., yield and product quality). 
Therefore, the digital twin should be \textit{calibrated} to reflect the physical system as accurately as possible, and it should be \textit{optimized} to identify an effective control policy for the physical system.

This paper considers a situation where the digital twin is specified by a finite number of \textit{calibration parameters}. The parameters are set based on experimental data from the physical system, which are expensive to collect and therefore very limited; often sample sizes are no more than 10-20 \citep{wang2024biomanufacturing}. However, under a particular parameter setting, the control policy obtained by simulating the digital twin can be used to guide the acquisition of new data \citep{powell2012optimal}. Thus, there is a synergy between the physical and digital systems, with the overall goal being to optimize the control policy in the real world. Both calibration and data acquisition must be linked back to policy performance in the physical system.
Therefore, digital twin calibration and informative data collection must be tailored to specific decision making of the real system:
(i) the prediction of the mean response for system design; and (ii) the dynamics characterized by a state-action transition model for system control.


Most existing studies on computer model calibration (see \cite{sung2024review} for a recent review) focus on statistical accuracy and uncertainty quantification for system design rather than control. Many of these studies build on the KOH framework, introduced by \cite{kennedy2001bayesian}, which uses Gaussian processes (GPs) to model the simulation output as well as the discrepancy from the physical system. \cite{efficient2015rui} refined the KOH framework to handle identifiability issues between calibration parameters and model discrepancy. Further improvements and extensions were developed by \cite{plumlee2016calibrating} and \cite{plumlee2017bayesian}. None of these papers considered the problem of control or modeled any policy optimization problem.

In marked contrast, we model both the physical and digital systems as infinite-horizon Markov decision processes (MDPs) with continuous state spaces. The goal in such problems is to learn an optimal policy function mapping a state to an action (or to a distribution over actions). The transition dynamics may have some known structure, but depend on parameters that are unknown for the physical system. The digital twin can be simulated with any choice of parameters, but since we do not know the ``true'' values, the simulation output may be an inaccurate representation of the real process \citep{luo2020learning}. Our proposed approach integrates parameter calibration with optimal control, strategically prioritizing experiments on the physical system to acquire information that yields the greatest benefit for policy optimization.

The problem of learning a state transition model characterizing the dynamics of a physical system, and then using this model to guide the search for a control policy, has been considered in the literature on model-based reinforcement learning (MBRL). See \cite{moerland2023model} for a survey describing various parametric and nonparametric state transition models. One widely used approach is based on GPs \citep{deisenroth2011pilco,
kurutach2018model}, which allow for a form of uncertainty quantification that can be explicitly considered when searching for a policy. However, it is quite difficult for GP-based MBRL to incorporate known structural information (for example, physics- and chemistry-based cell growth models). A different approach by \cite{wagenmaker2021task} imposes linear structure on the transition dynamics, but linearity is unable to capture the full complexity of many real-world systems. This reference paper also did not extensively study the connection between the data acquisition strategy, the model error, and the search for the optimal policy.

Our proposed ``Actor-Simulator'' framework uses a general nonlinear state transition model, which incorporates existing structural information (e.g., physics-based models of bioprocess dynamics) as well as uncertainty in the form of unknown parameters. Our approach alternates between calibrating the digital twin, identifying information to be acquired in the next experiment, and learning an reinforcement learning (RL) policy to optimize the physical system. A crucial component of this framework is a novel uncertainty function quantifying the discrepancy between the two systems with respect to the RL objective (i.e., the cumulative discounted reward). We use this criterion in two ways: during calibration, we select the next experiment to maximize the uncertainty function, and during policy optimization, we augment the reward function of the MDP by penalizing high-uncertainty actions. This approach leads to more effective targeted exploration of the state-action space, identifying those parts of it that yield the greatest benefit for policy optimization.

We present both theoretical and empirical evidence in favor of the new approach. First, on the theoretical side, we show that a consistent estimator of the model parameters leads to consistency of the policy. In other words, when calibration and optimization are linked through the Actor-Simulator, we eventually arrive at an optimal policy for the \textit{physical} system. We also discuss convergence rates. On the empirical side, we conduct extensive experiments on a problem motivated by biopharmaceutical manufacturing, with complex nonlinear dynamics and up to 40 unknown calibration parameters. The proposed approach significantly outperforms a state-of-the-art GP-based method as well as a baseline random policy.

In sum, this paper makes the following contributions. 
\begin{enumerate}
\item We propose a novel methodological framework linking a physical system to a digital twin model, in which calibration and optimization are conducted jointly. 
\item We create a novel uncertainty function quantifying the influence of model estimation uncertainty on the MBRL objective under general \textit{nonlinear} dynamics. 
\item We develop a novel calibration policy for information acquisition, targeting parts of the state-action space that are the most relevant for learning an optimal policy. 
\item We prove the convergence of the policy obtained from this approach. To the best of our knowledge, ours is the first theoretically justified MBRL method that does not rely on restrictive assumptions such as linear transitions.
\item We present extensive experimental results demonstrating that the proposed approach outperforms benchmarks, with respect to both parameter estimation and policy optimization, in a difficult and realistic application context.
\end{enumerate}


The rest of the paper is organized as follows. We review the most related studies in Section \ref{sec: related work} and present the problem description and outline the overall algorithmic framework for theoretical analysis in Section~\ref{sec: problem}. Then, we state the Actor-Simulator Algorithm in Section~\ref{sec: Algorithm}, prove its asymptotic optimality in Section~\ref{subsec: convergence analysis}, and present numerical experiments in Section \ref{sec: experiments}.
Finally, we draw conclusions in Section \ref{sec: conclusion}.


\section{Related Work} \label{sec: related work}

In this section, we review several related streams of literature from reinforcement learning, computer model calibration, and machine learning.


\noindent\textbf{Model Based Reinforcement Learning.} In reinforcement learning (RL), model-based approaches are widely considered to be more sample-efficient than model-free ones \citep{moerland2023model, nguyen2011model}. We may categorize this literature according to different types of process dynamics and state transition models used, for example, MBRL with linear MDPs \citep{jin2020provably}, MBRL with linear mixture models \citep{ayoub2020model,zhou2021nearly}, MBRL with Gaussian processes \citep{deisenroth2011pilco}, RL based on linear-quadratic regulators \citep{bradtke1992reinforcement,yang2019provably,wagenmaker2023optimal}, RL with neural networks \citep{levine2014learning,chua2018deep}, and ensemble-based RL \citep{kurutach2018model,Yu2020mopo,kidambi2020moreL}.  These approaches tend to impose too much or too little structure on the dynamics. In many situations, particularly in biobiopharmaceutical manufacturing, physics-based models provide more information than can be reflected in a GP or neural network, but such models cannot be forced into restrictive linear or linear-quadratic forms.

In the realm of empirical MBRL, a variety of model uncertainty-aware techniques have been developed, including approaches like uncertainty-aware RL 
\citep{chua2018deep,Yu2020mopo} 
as well as active exploration strategies outlined by \cite{nakka2020chance}. While these empirical methods have achieved notable success in practice, they often lack theoretical 
performance guarantees. The majority of theoretical research in this field has largely focused on more constrained settings, such as tabular models, Linear Quadratic Regulators (LQR), Gaussian Processes (GP), and linear MDPs. In addition, as \cite{song2021pc} pointed out, many existing MBRL algorithms lack the ability to perform
exploration and {thus cannot automatically adapt to exploration-challenging tasks}, especially dynamic decision making for complex stochastic systems such as biopharmaceutical manufacturing. {Our work mitigates these limitations by proposing a theoretically grounded calibration policy for systems with nonlinear dynamics, and by linking the exploration strategy to the policy optimization problem.}

\noindent\textbf{Computer Model Calibration.} 
Calibration methods have been widely used in biomanufacturing \citep{treloar2022deep, helleckes2022bayesian}, climate prediction \citep{higdon2013computer}, cosmology \citep{murphy2007methodology}, hydrology \citep{goh2013prediction}, cylinder implosion \citep{higdon2008computer}, fluidized-bed coating \citep{wang2009bayesian}, composite fuselages \citep{effective2020yan} and biomechanical engineering applications \citep{han2009simultaneous}. 
{Much of this literature uses the KOH framework of \cite{kennedy2001bayesian} as a foundation.} Recent work has started to combine calibration with optimization \citep{do2022metamodel}. \cite{li2019metamodel} introduce a metamodel-based approach, employing radial basis function metamodels to provide a smooth prediction surface for the objective value over the space of both design variables and preference parameters.
Similarly, \cite{chen2017efficient} advances a surrogate-based calibration methodology by seamlessly incorporating it into a simulation-optimization framework. 
{However, these papers generally consider static decision-making problems or system design in which the goal is to estimate a single response surface. Thus far, the literature on computer model calibration has not explored sequential decision-making problems with probabilistic state transitions.}

\noindent\textbf{Physics-Informed Machine Learning (ML) and Calibration.}
{Physics-Informed Machine Learning (PIML) is an emerging paradigm that integrates prior physical knowledge with data-driven machine learning models \citep{karniadakis2021physics}.  PIML is commonly used in biomanufaturing \citep{cui2024data}, fluid dynamics \citep{cai2021physics}, materials science \citep{zhang2020physics}, power systems \citep{misyris2020physics}, and biomechanical engineering applications \citep{cavanagh2021physics}. Calibration in PIML is enhanced by embedding physical laws into modeling, which guides mechanism learning, 
reduces the design space, and improves interpretability and sample-efficiency.
For example, to model the dynamics 
of Chinese Hamster Ovary (CHO) cell culture 
by using process data, \cite{cui2024data} integrate physical laws, such as mass balances and kinetic expressions for metabolic fluxes, into digital twin modeling. {However, this literature generally does not explicitly model an optimization problem, much less one where the decision variable is a control policy.}

\vspace{0.1in}
\section{Overview and Problem Description}
\label{sec: problem}

Section \ref{subsec: setting} presents the mathematical formalism used to model both the physical and digital systems. Section \ref{sec:framework} presents a high-level overview of our algorithmic framework and explains the role of each type of system in the procedure.


\subsection{Markov Decision Process Formulation}
\label{subsec: setting}

Throughout this paper, we model two related, but distinct stochastic systems: the \textit{physical system} and the \textit{digital twin}. Both systems involve sequential decisions made in an uncertain environment and they can be modeled using the language of Markov decision processes (MDPs). In this paper, we consider an infinite-horizon MDP represented by $\mathcal{M} = \left(\mathcal{S},\mathcal{A},r,\mu,\mu_0\right)$ that consists of a state space $\mathcal{S}$, an action space $\mathcal{A}$, a reward function $r:\mathcal{S}\times\mathcal{A} \rightarrow \mathbb{R}$, a transition probability 
{$\mu$} modeling the probability (or likelihood) $\mu\left(\pmb{s}'\mid \pmb{s},\pmb{a}\right)$ of being in state $\pmb{s}'\in\mathcal{S}$ after taking action $\pmb{a}\in\mathcal{A}$ out of state $\pmb{s}\in\mathcal{S}$, and a probability {density} $\mu_0$ representing the distribution of the initial state of the system. {For notational simplicity, we use $\mu(\cdot|\pmb{s},\pmb{a})$ and $\mu_0(\cdot)$ to represent the corresponding probability measure.}

Let $\mathcal{S},\mathcal{A}$ be compact subsets of Euclidean space, with $\sup_{\pmb{s}\in\mathcal{S}}\|\pmb{s}\|_2 \leq U^S$ and $\sup_{\pmb{a}\in\mathcal{A}}\|\pmb{a}\|_2 \leq U^A$ for some $U^S,U^A<\infty$. In biomanufacturing applications, state and action variables are continuous, but bounded, as biological systems have inherent physiological and engineering limits, such as temperature and pH, constrained by the capabilities and tolerances of biological systems (e.g., RNAs, cells, and proteins). 
We further suppose that ${0\leq}r\left(\pmb{s},\pmb{a}\right) \leq r^{\max}$ for some $r^{\max}<\infty$, as infinite rewards are not meaningful in practical applications. We use the notation $\mathcal{M}^p = \left(\mathcal{S},\mathcal{A},r,\mu^p,\mu_0\right)$ and $\mathcal{M}^d = \left(\mathcal{S},\mathcal{A},r,\mu^d,\mu_0\right)$ to refer to the physical system and digital twin, respectively. These MDPs use the same state and action spaces, reward function, and initial state distribution. They differ only in the transition probability kernel, reflecting the fact that the digital twin uses an imperfect model of the dynamics of the real-world system.

In biomanufacturing processes, the immediate effect of action $\pmb{a}$ (such as feeding strategy) on the state is often known. Then, by leveraging existing structural information on bioprocessing mechanisms, 
we impose a parametric structure on the state-transition model 
so that $\mu^p\left(\cdot\mid\pmb{s},\pmb{a}\right)$ can be written as $\mu\left(\cdot\mid\pmb{s},\pmb{a};\pmb{\beta}^\star\right)$, where the calibration parameters $\pmb{\beta}^\star$ in the biomanufacturing systems characterize the underlying reaction network regulation mechanisms, such as how cell growth, nutrient uptake, and metabolic waste generation rates depend on environmental conditions.
The digital twin will use the same parametric {function} $\mu$, but with an estimator $\hat{\pmb{\beta}}_n$ that is built on the historical data denoted by $\mathcal{D}_n$. Thus, the practitioner has some structural knowledge of the dynamics, represented by the function $\mu$. In the field of biochemistry, this knowledge typically comes from ``kinetic'' or ``mechanistic'' physics-based models \citep{Mandenius2013}. 
However, the true values $\pmb{\beta}^\star$ of the uncertain parameters are difficult to estimate, and the dynamics may be highly sensitive to their values. An important distinguishing feature of our work is that, aside from the presence of parameters, we do not impose additional structure on the dynamics. For example, we do not require transitions to be linear in the state variable \citep{simchowitz2018learning} or some transformation thereof \citep{wagenmaker2023optimal}.

A generic MDP $\mathcal{M}$ is controlled through a policy that takes in a state $\pmb{s}\in\mathcal{S}$ and chooses an action probabilistically. We denote by $\pi\left(\pmb{a}\mid\pmb{s}\right)$ the probability of choosing action $\pmb{a}$ out of state $\pmb{s}$ under policy $\pi$. The expected infinite-horizon discounted reward earned by a policy $\pi$ is denoted by
\begin{align}
 J\left(\pi;\mathcal{M}\right) \defeq \E^\pi_{\mu,\mu_0}\left[\sum^\infty_{n=0}\gamma^n r\left(\pmb{s}_n,\pmb{a}_n\right)\right] \label{eq: general objective}
\end{align}
where {$\gamma$ is the discount factor} and the expectation $\E^\pi_{\mu,\mu_0}$ is taken over the distribution $\mu_0$ of the initial state $\pmb{s}_0$ and over the conditional distributions $\mu\left(\pmb{s}_{n+1}\mid\pmb{s}_n,\pmb{a}_{n}\right)$ of each state $\pmb{s}_{n+1}$ given the previous state $\pmb{s}_n$, and under the assumption that $\pmb{a}_n$ is chosen with probability $\pi\left(\pmb{a}_n\mid\pmb{s}_n\right)$. It is well-known \citep{silver2014deterministic} that, if $\mathcal{M}$ is stationary and ergodic, $J\left(\pi;\mathcal{M}\right)$ admits the equivalent representation
\begin{equation*}
J\left(\pi;\mathcal{M}\right) = \E_{\left(\pmb{s},\pmb{a}\right)\sim d^\pi_{\mathcal{M}}\left(\pmb{s}\right)\pi\left(\pmb{a}\mid\pmb{s}\right)}[r(\pmb{s}, \pmb{a})],
\end{equation*}
with $d^{\pi}_{\mathcal{M}}(\pmb{s})=(1-\gamma)\int_{\mathcal{S}}
\sum_{t=0}^\infty\gamma^{t}\mu(\pmb{s}_{t}=\pmb{s}\mid\pmb{s}_0;{\pi}){\mu_0(\d\pmb{s}_0)}
$ denoting the limiting probability of state $\pmb{s}$ under policy $\pi$ and here we denote the density at state $\pmb{s}_t$ after transitioning for $t$
time steps from state $\pmb{s}_0$ by $\mu(\pmb{s}_{t}=\pmb{s}\mid\pmb{s}_0;{\pi})$. Similarly, we define the value function, i.e., 
\begin{align*}
 V^\pi\left(\pmb{s};\mathcal{M}\right) \defeq \E_{\mu}\left[\left. \sum^\infty_{n=0}\gamma^n r\left(\pmb{s}_n,\pmb{a}_n\right) \right| 
 \pmb{s}_0 = \pmb{s}\right],
\end{align*}
to be the value earned by policy $\pi$ when starting at a particular state $\pmb{s}$. 

Since each experiment is expensive, our \textit{objective} in this paper is to calibrate the digital twin model parameters $\pmb{\beta}$ to accelerate the search for the optimal policy 
$$\pi^\star \defeq \arg\max_{\pi\in\Pi} J\left(\pi;\mathcal{M}^p\right),$$ within some class $\Pi$, for the physical system $\mathcal{M}^p$.

\subsection{General Framework for Joint Calibration and Optimization}\label{sec:framework}

While the physical system and digital twin are closely related, the decision-maker interacts with them in very different ways. Interactions with the physical system take the form of experiments performed at selected state-action pairs $\left(\pmb{s},\pmb{a}\right)$, resulting in observations of the next state $\pmb{s}' \sim \mu\left(\cdot\mid\pmb{s},\pmb{a};\pmb{\beta}^\star\right)$. 
Therefore, the collection of experiments and their observed results is denoted by $\mathcal{D}_n=\{(\pmb{s}_i,\pmb{a}_i,\pmb{s}_{i+1})\}_{i=0}^n$, where $n$ is typically small because each experiment is typically expensive. Note that, in the context of MDPs, these data are neither independent nor identically distributed.

On the other hand, given some estimate of model parameter $\hat{\pmb{\beta}}$, the digital twin can simulate $\pmb{s}' \sim \mu(\cdot\mid\pmb{s},\pmb{a};\hat{\pmb{\beta}})$, which is much less expensive compared with the physical system. 
In fact, we can simulate an entire {trajectory} of states in a fraction of the time or cost that it would take to collect a single observation from the physical system. Given the wide array of reinforcement learning algorithms available in the literature, we may assume that it is relatively inexpensive to \textit{optimize} the MDP $\mathcal{M}^d$ under any particular estimated $\hat{\pmb{\beta}}$, or even with other modifications, e.g., to the reward function. In this way, we can obtain a policy that is optimal for this MDP, but not, of course, for the physical system. We therefore require a \textit{calibration} mechanism that uses the historical data $\mathcal{D}_n$ obtained from the physical system to configure the digital twin with some $\hat{\pmb{\beta}}_n$ so that we can efficiently reduce the digital twin prediction error measured by the uncertainty function $u$ (see more details in Section~\ref{sec: Algorithm}). 

To configure the digital twin $\mathcal{M}^d_n$ with some $\hat{\pmb{\beta}}_n$, letting $\hat{\pi}_n$ be the optimal policy of $\mathcal{M}^d_n$ (again, assuming that this is relatively easy to obtain), our overall goal is to ensure and accelerate the convergence of $\hat{\pi}_n$ to the optimal policy $\pi^\star$. An important aspect of our work is that each $\mathcal{M}^d_n$ will be used to guide the acquisition of new data.
 
The proposed Actor-Simulator approach works by iteratively calibrating the digital twin and is outlined in Algorithm~\ref{algo: actor simulator} that iteratively calibrates the digital twin and searches the optimal policy for the real system. Basically, in the $n$-th iteration, the algorithm performs three steps:
\begin{enumerate}[label=(\roman*)] 

\item \textbf{Digital Twin Calibration:} Given the state $\pmb{s}_{n-1}$ and the candidate policy $\hat{\pi}_{n-1}$ from the previous iteration, we select an action $\pmb{a}_{n-1}$ to maximize information gain, execute it in the physical system, and update 
$\mathcal{D}_n$ 
by adding $\left(\pmb{s}_{n-1},\pmb{a}_{n-1},\pmb{s}_{n}\right)$.

\item \textbf{Model Estimation:} We use the updated dataset $\mathcal{D}_{n}$ to build an estimate $\hat{\pmb{\beta}}_n$.

\item \textbf{Policy Optimization:} To account for the risk from model discrepancy, we construct the uncertainty-penalized reward $\Tilde{r}$ and the modified MDP $\Tilde{\mathcal{M}}_{n}^d$ with the estimates $\hat{\pmb{\beta}}_n$ and $\hat{\pi}_{n-1}$, and then solve it for the updated policy estimate $\hat{\pi}_n$.
\end{enumerate}

Note that the proposed Actor-Simulator algorithm closely integrates digital twin calibration and policy optimization.
{The policy $\hat{\pi}_{n}$ obtained in the policy optimization step is our estimate of $\pi^\star$, but it is also used in the digital twin calibration step in the selection of next action $\pmb{a}_{n}$, though the precise mechanism is more complicated than simply choosing the action prescribed by the policy. We see, however, that the two systems work together: data from the physical system are used to calibrate the digital twin, and the updated policy from the digital twin is used to guide the collection of new data.}
We call this framework ``Actor-Simulator,'' inspired by actor-critic methods from the reinforcement learning literature \citep{konda2003onactor}. The key difference is that those methods worked only with a physical system, and updated a policy iteratively after each action. In our setting, the presence of a digital twin allows us to learn an optimal policy for an MDP very easily; the difficulty is that this policy is optimal only for an approximation of the physical system.

A more formal statement of the Actor-Simulator procedure is given in Algorithm~\ref{algo: actor simulator}. It can be seen that both the calibration and optimization steps depend on an uncertainty function $u$. The formal definition of this criterion is deferred to Section~\ref{sec: Algorithm},
but in brief, it is an upper bound on the discrepancy between the predicted value of $\hat{\pi}_n$ under the true and approximate dynamics, respectively. In other words, it quantifies the error made by the digital twin in computing the value of the current policy. As $\hat{\pmb{\beta}}_n$ becomes more accurate, this discrepancy is reduced.

Calibration and optimization use the uncertainty function in very different ways. When we perform calibration, we interact directly with the physical system, giving us an opportunity to collect new data and improve the accuracy of the digital twin. We therefore look for actions with high uncertainty, because learning about these will tend to yield the greatest information gain and reduction in $u$. Thus, step~1 in Algorithm~\ref{algo: actor simulator} is risk-seeking, preferring higher uncertainty.

On the other hand, when we perform policy optimization, we are working only with the digital twin, and do not interact with the physical system at all. In this step, we are not able to collect any real data to improve our estimates. In such a situation, maximizing $u$ will have the effect of amplifying the inaccuracies in the digital twin model, leading to erratic and unstable results. Instead, in steps 5-6 of Algorithm \ref{algo: actor simulator}, we use $u$ to \textit{penalize} the reward of $\mathcal{M}^d_n$, leading to risk-averse policies. 
The distinction between risk-seeking calibration and risk-averse optimization is intentional and important for practical performance.

\begin{algorithm}[ht!]
\SetAlgoLined
\DontPrintSemicolon
\hspace{0.05em}\textbf{Input}: $K$ maximum number of iterations, uncertainty penalization parameter $\lambda$, reward function $r(\cdot,\cdot)$, 
and initial measure $\mu_0$\;
\textbf{Output}: calibrated dynamics $\hat{\pmb{\beta}}_K$ and the optimal policy $\hat\pi_K$\; 
\textbf{Initialize} parameters of digital twin $\hat{\pmb{\beta}}_0$, initial state $\pmb{s}_0\sim \mu_0(\cdot)$, and $\mathcal{D}_0= \varnothing$\;
 \hspace{0.05em}\For{$n=1,2,\ldots, K$}{
    \hspace{1em}\textbf{(i) Digital Twin Calibration:} \\
    \quad 1. Given state $\pmb{s}_{n-1} \in\mathcal{S}$ and parameter $\hat{{\pmb{\beta}}}_{n-1}$, compute the action\;
     \hspace{2.1em}$\pmb{a}_{n-1} = \argmax\limits_{\pmb{a}\in\mathcal{A}} u(\pmb{s}_{n-1},\pmb{a};\hat{{\pmb{\beta}}}_{n-1}, \hat\pi_{n-1})$;\;
    \hspace{1em}2. Observe the next state $\pmb{s}_{n}\sim {\mu}(\pmb{s}_{n}|\pmb{s}_{n-1},\pmb{a}_{n-1}; \pmb{\beta}^\star)$ (physical system);\;
      \hspace{1em}3. Update the historical dataset $\mathcal{D}_n=\mathcal{D}_{n-1} \cup \{(\pmb{s}_{n-1}, \pmb{a}_{n-1}, \pmb{s}_n)\}$;\;
     \hspace{1em}\textbf{(ii) Model Estimation:}\\
    \quad 4. Use the dataset $\mathcal{D}_n$ to update the model parameter estimate $\hat{{\pmb{\beta}}}_n$;\;
     \hspace{1em}\textbf{(iii) Policy Optimization:}\\ 
    \quad 5. Compute the uncertainty-penalized reward\;
    \hspace{2.1em}$\tilde{r}(\pmb{s},\pmb{a})\defeq r(\pmb{s},\pmb{a})-\lambda u(\pmb{s},\pmb{a}; \hat{\pmb{\beta}}_n,\hat\pi_{n-1})$ and the modified MDP\;
    \hspace{2.1em}$\Tilde{\mathcal{M}}_{n}^d=(\mathcal{S}, \mathcal{A}, \Tilde{r},{\mu}(\cdot;\hat{\pmb{\beta}}_n),\mu_0)$;\;
    \hspace{1em}6. 
   {Use a RL algorithm} to solve $\Tilde{\mathcal{M}}_{n}^d$ and return the updated policy $\hat\pi_{n}$.
   
 }
\caption{Actor-Simulator Algorithm}\label{algo: actor simulator}
\end{algorithm}

\section{The Actor-Simulator: Algorithmic Overview}
\label{sec: Algorithm}

As described in Section \ref{sec:framework}, the proposed Actor-Simulator framework consists of three key steps. Sections~\ref{subsec: calibration setup}-\ref{sec:policyOptimizationstep} give precise specifications of model estimation, calibration, and actor update for policy optimization.
We also present theoretical properties of each step that will eventually be relevant with the convergence analysis in Section~\ref{subsec: convergence analysis}.

\subsection{Model Estimation} 
\label{subsec: calibration setup}

Suppose that we are given the dataset $\mathcal{D}_n$ collected from the physical system.
We use maximum likelihood estimation (MLE), which minimizes the Kullback-Leibler divergence or, equivalently, maximizes the log-likelihood denoted by $\ell(\cdot)$, i.e.,
$$\hat{\pmb\beta}=\argmin_{{\pmb{\beta}}} D_{KL}\left(\mu(\pmb{s}^\prime|\pmb{s},\pmb{a};{\pmb{\beta}}^\star\right)\Vert \mu(\pmb{s}^\prime|\pmb{s},\pmb{a};{\pmb{\beta}}))=\argmax_{{\pmb{\beta}}} \E[\ell(\pmb{s},\pmb{a},\pmb{s}^\prime;{\pmb{\beta}})].$$
In practice, one uses the empirical expectation, giving rise to the problem
\begin{equation}\label{eq: calibration estimator}
\hat{{\pmb{\beta}}}_n\defeq \argmax_{{\pmb{\beta}}}\frac{1}{n}\sum^n_{i=1}\ell(\pmb{s}_i,\pmb{a}_i,\pmb{s}_{i+1};{\pmb{\beta}})
\end{equation}
{where $\ell(\pmb{s}_i,\pmb{a}_i,\pmb{s}_{i+1};{\pmb{\beta}}) \defeq
\log {\mu}(\pmb{s}_{i+1}|\pmb{s}_i,\pmb{a}_i;{\pmb{\beta}})$ is the log-likelihood.}
It is worth noting that the $L_2$ distance used by \cite{tuo2015efficient} is a special case of \eqref{eq: calibration estimator} under Gaussian assumptions on the underlying distribution.


The consistency of $\hat{\pmb{\beta}}_n$ is not a foregone conclusion, because the data in $\mathcal{D}_n$ do not obey standard independent assumptions. For dependent observations, strong consistency and asymptotic normality have been investigated under various regularity conditions and assumptions on the nature of the dependence \citep{crowder1976maximum,heijmans1986first,tinkl2013asymptotic}. In this paper, the model estimation builds on the results of \cite{tinkl2013asymptotic}, that establishes the asymptotic normality of the MLE under the assumption that the sample path is a strictly stationary and ergodic $\alpha$-mixing process (Definition~\ref{def:mixing-process}). 

\begin{definition}[\citealp{bradley2005basic}]\label{def:mixing-process}
Suppose {$\{Y_i\}_{i\in\mathbb{Z}} = \{(\pmb{s}_i,\pmb{a}_i,\pmb{s}_{i+1})\}_{i\in\mathbb{Z}}$} is a strictly stationary sequence of random vectors and ergodic for any
policy $\pi\in\Pi$. The $\alpha$-mixing coefficient $\alpha(T)$ 
at time $T$ is defined as:
\begin{align*}
   {\alpha(T)}&\defeq
   \sup_{A\in \mathcal{F}^0_{-\infty},B\in  \mathcal{F}^{\infty}_T} \left|P(A\cap B) - P(A)P(B)\right|
\end{align*}
where $\mathcal{F}^0_{-\infty}$ and $\mathcal{F}^{\infty}_T$ represent the sigma-algebra generated by $\{Y_i: i\leq 0\}$ and $\{Y_i: i \geq T\}$ respectively.
\end{definition}

We first present the necessary regularity conditions in Assumption \ref{assumption 4}, followed by the main asymptotic result in Lemma~\ref{lemma: asymptotic convergence}. 


\begin{assumption}[Regularity Conditions]\label{Regularity Cond}
\quad\\
The following conditions hold:
 \begin{enumerate}[label=(\alph*)]
     \item The $\alpha$-mixing process $\{(\pmb{s}_i,\pmb{a}_i, \pmb{s}_{i+1})\}_{i\in\mathbb{Z}}$ is strictly stationary and ergodic for any policy $\pi\in\Pi$.
     \item The log-likelihood function 
     $\ell(\pmb{s},\pmb{a},\pmb{s}^\prime;{\pmb{\beta}})$ is continuous in model parameters ${\pmb{\beta}} \in \mathbb{B}$ for all $(\pmb{s},\pmb{a},\pmb{s}^\prime)$ and $\mathbb{B}\subset \mathbb{R}^d$ is compact.
     \item The expected log-likelihood is bounded, $\E[\ell(\pmb{s},\pmb{a},\pmb{s}^\prime;{\pmb{\beta}}^\star)] <\infty$ for all $(\pmb{s},\pmb{a},\pmb{s}^\prime)$. 
     \item {For every compact set $G$ containing ${\pmb{\beta}}^\star$, it holds $\mathcal{L}({\pmb{\beta}}^\star)
     > 
     \mathop{\sup}\limits_{{\pmb{\beta}}\notin G} \mathcal{L}({\pmb{\beta}})$ and also $$\mathop{\lim\sup}_{n\rightarrow\infty}\mathop{\sup}_{{\pmb{\beta}} \in\mathbb{B}} \frac{1}{n}\sum^n_{i=1}\ell(\pmb{s}_i,\pmb{a}_i,\pmb{s}_{i+1};{\pmb{\beta}})\leq\mathop{\sup}_{{\pmb{\beta}} \in\mathbb{B}}  \mathcal{L}({\pmb{\beta}})$$ almost surely (a.s.), where the expected log-likelihood function is defined as,
     \begin{equation}
     \mathcal{L}({\pmb{\beta}})
    \defeq
     \E_{\pmb{s}^\prime\sim {\mu}(\cdot|\pmb{s},\pmb{a};{\pmb{\beta}}^\star)d^\pi_{\mathcal{M}^p}(\pmb{s},\pmb{a})}[\ell(\pmb{s},\pmb{a},\pmb{s}^\prime;{\pmb{\beta}})].
     \label{eq.expectedLoss}
     \end{equation}}
     \item $\ell(\pmb{s},\pmb{a},\pmb{s}^\prime;{\pmb{\beta}})$ is continuously differentable w.r.t. ${\pmb{\beta}}$ almost everywhere. For the derivative of log-likelihood function $\nabla\ell(\pmb{s},\pmb{a},\pmb{s}^\prime;{\pmb{\beta}})$, there exists an integrable function $L(\pmb{s},\pmb{a},\pmb{s}^\prime)$ such that $\Vert \nabla\ell(\pmb{s},\pmb{a},\pmb{s}^\prime;{\pmb{\beta}})\Vert\leq L(\pmb{s},\pmb{a},\pmb{s}^\prime)$ in a neighborhood of ${\pmb{\beta}}^\star$.
     \item The derivative $\nabla\ell(\pmb{s},\pmb{a},\pmb{s}^\prime;{\pmb{\beta}})$ is measureable for all ${\pmb{\beta}}$ and differentiable w.r.t. ${\pmb{\beta}}$ almost everywhere. For the second-order derivative $\nabla^2\ell(\pmb{s},\pmb{a},\pmb{s}^\prime;{\pmb{\beta}})$, there exists an integrable function $U(\pmb{s},\pmb{a},\pmb{s}^\prime)$ such that $\Vert \nabla^2\ell(\pmb{s},\pmb{a},\pmb{s}^\prime;{\pmb{\beta}}) \Vert\leq U(\pmb{s},\pmb{a},\pmb{s}^\prime)$ in a neighborhood of ${\pmb{\beta}}^\star$.
     \item {$\E[\nabla^2\ell(\pmb{s},\pmb{a},\pmb{s}^\prime;{\pmb{\beta}})]$ is negative definite for any $\pmb\beta$.}
 \end{enumerate}
 \label{assumption 4}
\end{assumption}
{These regularity conditions are collected from \cite{tinkl2013asymptotic}, Corollary 4.3.12 and \cite{vandervaart1996weak}, Lemma 3.2.1.} For the asymptotic convergence analysis in Section~\ref{subsec: convergence analysis}, we only require mild conditions (Assumption~\ref{assumption 4}(\textit{a}-\textit{d})).
In contrast, the additional conditions (e-g) are required for establishing the asymptotic normality in Lemma~\ref{lemma: asymptotic convergence} and used in approximating the uncertainty function in Section~\ref{sec:simulationstep}.


 \begin{lemma}[\citealp{tinkl2013asymptotic}, Corollary 4.3.12] 
 \label{lemma: asymptotic convergence} 
 Let the trajectory $\{(\pmb{s}_i,\pmb{a}_i,\pmb{s}_{i+1})\}_{i\in \mathbb{Z}}$ satisfies Assumption~\ref{assumption 4}(a-d). Then for the estimator $\hat{{\pmb{\beta}}}_n$ defined
 by \eqref{eq: calibration estimator}, we have
 $$\hat{{\pmb{\beta}}}_n\xrightarrow{a.s.} {\pmb{\beta}}^\star.$$
In addition, if {Assumption~\ref{assumption 4}(e-g)} are met, we further have
     $$\sqrt{n}(\hat{{\pmb{\beta}}}_n - {\pmb{\beta}}^\star) \xrightarrow{d} \mathcal{N}(0,\Sigma({\pmb{\beta}}^\star))$$
where $\Sigma({\pmb{\beta}}^\star) \defeq - \E[\nabla^2 \ell(\pmb{s},\pmb{a},\pmb{s}^\prime;{\pmb{\beta}}^\star)^\top]^{-1}$ is the inverse of Fisher information matrix.
\label{lemma: asymptotic normality for dependent data}
 \end{lemma}

We note that, in general, the asymptotic behavior of $\hat{\pmb{\beta}}_n$ is challenging to characterize. Virtually all results of this type assume some particular structure on the sample dependence. Lemma~\ref{lemma: asymptotic convergence} is not the only possible result. For example, by assuming $\phi$-mixing,
\cite{crowder1976maximum} presented a different rate of convergence based on the sequence of smallest eigenvalues of Fisher information matrices.

The $\alpha$-mixing condition (Definition~\ref{def:mixing-process}) is less restrictive than what has been assumed in the recent RL literature. Notably, recent theoretical analyses of RL algorithms often rely on a stronger geometric ergodicity condition \citep{bhandari2018finite, xu2020improving, wu2020finite, zheng2022variance}, which requires the transition probability distribution to converge to a stationary distribution at an exponentially fast rate. Under the $\alpha$-mixing assumption, 
we can obtain additional results on the tail behavior of the conditional expected {log-likelihood},
\begin{equation*}
\mathcal{L}\left(\hat{\pmb{\beta}}\right) 
= 
\E\left[
\left. \ell\left(\pmb{s},\pmb{a},\pmb{s}^\prime;{\hat{\pmb{\beta}}}\right)
\right|
\hat{\pmb{\beta}}\right].
\end{equation*}
The following result presented in Lemma \ref{lemma: uniformly tight} is 
obtained by applying the delta method to the asymptotic normality in Lemma \ref{lemma: asymptotic convergence}
and invoking 
Theorem 2.4 (Prohorov's theorem) in \cite{van2000asymptotic}.

\begin{lemma}\label{lemma: uniformly tight}
Let $X_n \defeq \sqrt{n}({\mathcal{L}(\hat{\pmb{\beta}}}_n) - \mathcal{L}({\pmb{\beta}}^\star))$. If Assumption \ref{assumption 4} holds, then $\{X_n\}$ is uniformly bounded in probability, i.e., for every $\delta>0$, there exists a large deviation $M_{n,\delta}>0$ such that
\begin{equation*}
\sup_{n\geq 1} P(\Vert X_n\Vert > M_{n,\delta})<\delta, 
\end{equation*}
where $M_{n,\delta}$ depends on both sample size $n$ and tail probability $\delta$.
\end{lemma}

Using Lemma \ref{lemma: uniformly tight}, we can also show a convergence rate as a function of $M_{n,\delta}$ for the expected loss in Theorem~\ref{thm:expectedlossrate}. The proof is given in {Appendix \ref{exploss}}.

\begin{theorem}\label{thm:expectedlossrate}
Suppose that we are in the situation of Lemma \ref{lemma: uniformly tight}. Then,
\begin{equation*}
\E\left[\mathcal{L}\left(\hat{\pmb{\beta}}_n\right) - \mathcal{L}\left(\pmb{\beta}^\star\right)\right] 
=
\mathcal{O}\left(\max\left\{\frac{M_{n,\delta}}{\sqrt{n}},\delta\right\}\right).
\end{equation*}
\end{theorem}

{The precise determination of $\delta$ and $M_{n,\delta}$ depends on the underlying physical system and is outside the scope of this paper. A smaller $M_{n,\delta}$ indicates greater confidence in the parameter estimates and, consequently, a more accurate and reliable model.} Table~\ref{table: high probability bound review} provides an overview of known bounds for various settings in the literature. However, these results assume either linear transition dynamics or independent samples, illustrating the difficulty of showing explicit results in the general case. 

\begin{table}[t]
\caption{High probability bounds for physical systems under various assumptions.}\label{table: high probability bound review}
\normalsize
\centering
\fontsize{8pt}{12pt}\selectfont
\addtolength{\tabcolsep}{-0.4em}
\begin{tabular}{@{}c|ccccc@{}}
\toprule
Physical System                                                            & Key Assumptions   & High Probability Bound                                                                   & $M_{n,\delta}$                                             & \begin{tabular}[c]{@{}c@{}} {Sample}\\ Dependence\end{tabular}   \\ \midrule
\begin{tabular}[c]{@{}c@{}}Linear\\ \citep{simchowitz2018learning}\end{tabular}
   & \begin{tabular}[c]{@{}c@{}}Martingale Small-Ball\\ sub-Gaussian Noise\end{tabular}                  & $\mbox{Pr}\left(\Vert\hat{\pmb{\beta}}_n-{\pmb{\beta}}^\star\Vert >\frac{M_{n,\delta}}{\sqrt{n}}\right)\leq 3\delta$                & $\mathcal{O}\left(\sqrt{\log\left(1/\delta\right)}\right)$                                             & Dependent         \\
  \begin{tabular}[c]{@{}c@{}}Nonlinear\\ \citep{hazan2014beyond}\end{tabular}                      & \begin{tabular}[c]{@{}c@{}}Strong Convexity\\ Bounded Gradient\end{tabular} & $\mbox{Pr}\left(\mathcal{L}(\hat{\pmb{\beta}}_n)-\mathcal{L}({\pmb{\beta}}^\star)>\frac{M_{n,\delta}}{\sqrt{n}}\right)\leq \delta$ & $\mathcal{O}\left(\frac{\log(1/\delta)+\log\log(n)}{\sqrt{n}}\right)$ & Independent       \\
\begin{tabular}[c]{@{}c@{}}Nonlinear\\ \citep{harvey2019tight}\end{tabular}        &   \begin{tabular}[c]{@{}c@{}}Strong Convexity\\ Lipschitz Continuity\end{tabular}                                 &    $\mbox{Pr}\left(\mathcal{L}(\hat{\pmb{\beta}}_n)-\mathcal{L}({\pmb{\beta}}^\star)>\frac{M_{n,\delta}}{\sqrt{n}}\right)\leq \delta$                                                                                         & $\mathcal{O}\left(\frac{\log(n)\log(1/\delta)}{\sqrt{n}}\right)$                                                           & Independent                  \\
\begin{tabular}[c]{@{}c@{}}Nonlinear\\ \citep{harvey2019tight}\end{tabular}      &   \begin{tabular}[c]{@{}c@{}}Strong Convexity\\ Lipschitz Continuity\end{tabular}                                 &     $\mbox{Pr}\left(\mathcal{L}(\hat{\pmb{\beta}}_n)-\mathcal{L}({\pmb{\beta}}^\star)>\frac{M_{n,\delta}}{\sqrt{n}}\right)\leq \delta$                                                                                         & $\mathcal{O}\left({\log(n)\log(1/\delta)}\right)$                                                           & Independent                  \\
\bottomrule
\end{tabular}
\end{table}

\subsection{Digital Twin Calibration}
\label{sec:simulationstep}

Given any candidate policy $\pi$, when the digital twin is used to predict the performance of the physical system 
$J\left(\pi; \mathcal{M}^p\right)$, 
the main issue is to relate the calibration error, i.e., the error in the estimated parameters $\hat{\pmb{\beta}}$, to the error in the estimate of the RL objective function made by the digital twin. The following result from \cite{Yu2020mopo} makes this relationship explicit.

\begin{lemma}[\citealp{Yu2020mopo}, Theorem~4.4]\label{lemma: objective gap between MDPs}
Let $\mathcal{M}^p$ and $\mathcal{M}^d$ be two MDPs with the same reward function $r$, but different dynamics $\mu^p$ and $\mu^d$, respectively. For any policy $\pi\in\Pi$, let $G^\pi(\pmb{s},\pmb{a})\defeq 
 \E_{\pmb{s}^\prime\sim\mu^d(\cdot|\pmb{s},\pmb{a})}\left[V^\pi\left(\pmb{s}^\prime;\mathcal{M}^p\right)\right]-\E_{\pmb{s}^\prime\sim{{\mu^p}}(\cdot|\pmb{s},\pmb{a})}\left[V^\pi(\pmb{s}^\prime;\mathcal{M}^p)\right]$. Then,
     $$J\left(\pi; \mathcal{M}^d\right)-J\left(\pi; \mathcal{M}^p\right)
     =\gamma
     \E_{(\pmb{s},\pmb{a})\sim d_{\mathcal{M}^d}^{\pi}(\pmb{s})\pi(\pmb{a}|\pmb{s})}\left[G^\pi(\pmb{s},\pmb{a})\right].$$
\end{lemma}
Lemma~\ref{lemma: objective gap between MDPs} establishes the calibration discrepancy between the digital twin and the physical system through the quantity $\mathop{\E}_{(\pmb{s},\pmb{a})\sim d_{\mathcal{M}^d}^{\pi}(s)\pi(\pmb{a}|\pmb{s})}\left[G^\pi(\pmb{s},\pmb{a})\right]$. However, this quantity depends on the value function of the physical system and is difficult to compute. {To address this issue, we apply Theorem 2.1 from \cite{bolley2005weighted} to propose a more tractable upper bound on $G^\pi(\pmb{s},\pmb{a})$ in Theorem~\ref{thm: uncertainty}.}


\begin{theorem}\label{thm: uncertainty}
Suppose we are given a policy $\pi \in \Pi$ and an estimate $\hat{\pmb{\beta}}$. Define the uncertainty function of the state-action pair $(\pmb{s},\pmb{a})\in \mathcal{S}\times \mathcal{A}$ as
 \begin{align}\label{eq:udef}
     &u(\pmb{s},\pmb{a};\hat{{\pmb{\beta}}},\pi)
     \defeq \sqrt{2}\left(1+\log \E_{\mu\left(\pmb{s}^\prime|\pmb{s},\pmb{a};\hat{\pmb{\beta}}\right)}\left[e^{V^\pi(\pmb{s}^\prime;\mathcal{M}^p)^2}\right]\right)^{{1/2}}\sqrt{D_{KL}\left(\mu(\cdot|\pmb{s},\pmb{a}; {\pmb{\beta}}^\star)\Vert \mu(\cdot|\pmb{s},\pmb{a}; \hat{\pmb{\beta}})\right)}.
 \end{align}
Then, we have
\begin{equation}\label{eq: model estimation error}
   G^\pi(\pmb{s},\pmb{a}) \leq u(\pmb{s},\pmb{a};\hat{{\pmb{\beta}}},\pi)
\end{equation}
where $G^\pi(\pmb{s},\pmb{a})$ is as in Lemma~\ref{lemma: objective gap between MDPs}.
\end{theorem}

The uncertainty function $u$ incorporates the estimation error of the state transition model, represented by $D_{KL}\left(\mu(\cdot|\pmb{s},\pmb{a}; {\pmb{\beta}}^\star)\Vert \mu(\cdot|\pmb{s},\pmb{a}; \hat{\pmb{\beta}})\right)$, and a weight $1+\log \E_{\mu\left(\pmb{s}^\prime|\pmb{s},\pmb{a};\hat{\pmb{\beta}}\right)}\left[e^{V^\pi(\pmb{s}^\prime;\mathcal{M}^p)^2}\right]$. The first term measures the pointwise discrepancy between the state transition models of the digital twin and the physical system. The second term relates the uncertainty quantification to the objective value of the policy by including the value function. Combining Lemma \ref{lemma: objective gap between MDPs} and Theorem \ref{thm: uncertainty} yields
 \begin{equation}\label{eq:Jbound}
     |J(\pi;\mathcal{M}^d) - J(\pi;\mathcal{M}^p)| 
     \leq {\gamma} 
     \E_{(\pmb{s},\pmb{a})\sim d_{\mathcal{M}^d}^{\pi}(\pmb{s})\pi(\pmb{a}|\pmb{s})}\left[u(\pmb{s},\pmb{a};\hat{\pmb{\beta}},\pi)\right].
 \end{equation}
 
Eq.~(\ref{eq:Jbound}) shows 
how the uncertainty function $u(\pmb{s},\pmb{a};\hat{\pmb{\beta}},\pi)$ connects the physical and digital systems. Recall from Algorithm~\ref{algo: actor simulator} that $u$ is used in both digital twin calibration and policy optimization. We will not repeat the details here; instead, in the remainder of this subsection, we will focus on the computation of $u$. 

From (\ref{eq:udef}), we see that $u$ depends on $\mathcal{M}^p$, and therefore cannot be computed exactly. Thus, we develop an approximation as follows. Given $(\pmb{s}, \pmb{a}, \hat{\pmb{\beta}}_{n}, \hat{\pi}_{n})$, we denote
\begin{equation}\label{eq:defofw}
{w(\pmb{s},\pmb{a};\hat{\pmb{\beta}}_{n}, \hat{\pi}_{n})=2\left(1+\log \E_{\mu\left(\pmb{s}^\prime|\pmb{s},\pmb{a};{\hat{\pmb{\beta}}_{n}}\right)}\left[e^{V^{\hat{\pi}_{n}}(\pmb{s}^\prime;\mathcal{M}^p)^2}\right]\right)}
\end{equation}
for convenience. Observe that, by the definition of the KL divergence,
\begin{eqnarray}
   u^2\left(\pmb{s},\pmb{a};\hat{\pmb{\beta}}_{n},\hat{\pi}_{n}\right)  
   =w\left(\pmb{s},\pmb{a};\hat{\pmb{\beta}}_{n}, \hat{\pi}_{n}\right)
   \E_{\mu^p}\left[\log {\mu}(\pmb{s}^\prime|\pmb{s},\pmb{a};{\pmb{\beta}}^\star)-\log {\mu}(\pmb{s}^\prime|\pmb{s},\pmb{a};\hat{\pmb{\beta}}_{n})\right]. 
   \label{eq: exploration policy}
\end{eqnarray}
We then take a second-order Taylor expansion of $\E_{\mu^p}[\log {\mu}(\pmb{s}^\prime|\pmb{s},\pmb{a};\hat{\pmb{\beta}}_n)]$ at ${\pmb{\beta}}^\star$. Since
\begin{equation*}
\E_{\mu^p}[\nabla \log {\mu}(\pmb{s}^\prime|\pmb{s},\pmb{a};{\pmb{\beta}}^\star)]=\int \nabla {\mu}(\pmb{s}^\prime|\pmb{s},\pmb{a};{\pmb{\beta}}^\star)\d\pmb{s}^\prime=\nabla\int {\mu}(\pmb{s}^\prime|\pmb{s},\pmb{a};{\pmb{\beta}}^\star)\d\pmb{s}^\prime=0,
\end{equation*}
we have
\begin{eqnarray}
\E_{\mu^p}[\log {\mu}(\pmb{s}^\prime|\pmb{s},\pmb{a};\hat{\pmb{\beta}}_n)]
  \approx \E_{\mu^p}\left[\log {\mu}(\pmb{s}^\prime|\pmb{s},\pmb{a};{\pmb{\beta}}^\star)\right]
    +\frac{1}{2} (\hat{\pmb{\beta}}_n-{\pmb{\beta}}^\star)^\top\E_{\mu^p}\left[\nabla^2 \log {\mu}(\pmb{s}^\prime|\pmb{s},\pmb{a};{\pmb{\beta}}^\star)\right](\hat{\pmb{\beta}}_n-{\pmb{\beta}}^\star). \label{eq: approximation of log likelihood}
\end{eqnarray}
By substituting 
$\E[\log {\mu}(\pmb{s}^\prime|\pmb{s},\pmb{a};\hat{\pmb{\beta}}_n)]$ in \eqref{eq: exploration policy} with its approximation \eqref{eq: approximation of log likelihood}, we have
\begin{equation}
u^2\left(\pmb{s},\pmb{a};\hat{{\pmb{\beta}}}_n,\hat{\pi}_{n}\right) \approx w\left(\pmb{s},\pmb{a};\hat{\pmb{\beta}}_n, \hat{\pi}_{n}\right) \Vert\hat{\pmb{\beta}}_n-{\pmb{\beta}}^\star\Vert^2_{\mathcal{I}({\pmb{\beta}}^\star;\pmb{s},\pmb{a})},\label{eq 2: exploration policy}
\end{equation}
where the \textit{conditional} Fisher information matrix, denoted by $$\mathcal{I}({\pmb{\beta}}^\star;\pmb{s},\pmb{a})
\defeq
- \E_{\mu^p}\left[\nabla^2 \log {\mu}(\pmb{s}^\prime|\pmb{s},\pmb{a};{\pmb{\beta}}^\star)\right],$$ 
is used to design the calibration criteria $u$ that can guide the selection of next action $\pmb{a}$ for given current state $\pmb{s}$, maximizing information gain. The term $\Vert\hat{\pmb{\beta}}_n-{\pmb{\beta}}^\star\Vert^2_{\mathcal{I}({\pmb{\beta}}^\star;\pmb{s},\pmb{a})}$ is a Fisher information matrix weighted Euclidean distance that measures the discrepancy between the estimator $\hat{\pmb{\beta}}_n$ and the true value
${\pmb{\beta}}^\star$, taking into account the variability and correlation of the estimates as captured by the information matrix. This essentially scales each component of the difference $\hat{\pmb{\beta}}_n-{\pmb{\beta}}^\star$ according to the information gain. We assume that $\mathcal{I}({\pmb{\beta}}^\star;\pmb{s},\pmb{a})$ is nonsingular unless specified otherwise; such assumptions are common in the literature on experimental design for nonlinear models {\citep{yang2013optimal}}.

Due to the asymptotic normality of $\hat{\pmb{\beta}}_n$ as shown in Lemma~\ref{lemma: asymptotic convergence}, we may apply the continuous mapping theorem to obtain $$\Vert\hat{\pmb{\beta}}_n-{\pmb{\beta}}^\star\Vert^2_{\mathcal{I}({\pmb{\beta}}^\star;\pmb{s},\pmb{a})}\xrightarrow{d} G(v)$$
where $G(v)$ is a weighted chi-square distribution \citep{solomon1977distribution} with $v=(v_1,v_2,\ldots,v_d)$ denoting the eigenvalues of $\mathcal{I}({\pmb{\beta}}^\star;\pmb{s},\pmb{a})\Sigma({\pmb{\beta}}^\star)$, where $\Sigma({\pmb{\beta}}^\star)$ is the asymptotic covariance matrix obtained from Lemma \ref{lemma: asymptotic convergence}. From the properties of this distribution,
\begin{equation}\label{eq: weighted distance estimation}
   \E\left[G\left(v\right)\right]=\sum^d_{i=1}v_i=\Tr\left(\mathcal{I}({\pmb{\beta}}^\star;\pmb{s},\pmb{a})\Sigma({\pmb{\beta}}^\star)\right).
\end{equation}
We now simply use plug-in estimators to write
\begin{equation}
u^2\left(\pmb{s},\pmb{a};\hat{{\pmb{\beta}}}_{n},\hat{\pi}_{n}\right) \approx \hat{w}\left(\pmb{s},\pmb{a};\hat{\pmb{\beta}}_{n}, \hat{\pi}_{n}\right) \Tr({\mathcal{I}(\hat{\pmb{\beta}}_{n};\pmb{s},\pmb{a})}\Sigma(\hat{\pmb{\beta}}_{n}))\label{eq: practical exploration policy}
\end{equation}
where
\begin{equation*}
{\hat{w}(\pmb{s},\pmb{a};{\hat{\pmb{\beta}}}_{n}, \hat{\pi}_{n})=2\left(1+\log \E_{\mu(\pmb{s}^\prime|\pmb{s},\pmb{a};\hat{\pmb{\beta}}_{n})}\left[e^{V^{\hat{\pi}_{n}}(\pmb{s}^\prime;\mathcal{M}^d_{n})^2}\right]\right)}
\end{equation*}
uses the digital twin $\mathcal{M}^d_{n}$ from the previous iteration instead of $\mathcal{M}^p$.

In practice, the asymptotic covariance can be estimated by using the inverse of the Hessian matrix, $\Sigma(\hat{\pmb{\beta}}_n)^{-1} = - \frac{1}{n}\sum^n_{i=1}\nabla^2\ell(\pmb{s}_i,\pmb{a}_i,\pmb{s}_{i+1};\hat{\pmb{\beta}}_n)$. 
{The estimation of the conditional Fisher information matrix, i.e., $\mathcal{I}({\hat{\pmb{\beta}}_n};\pmb{s},\pmb{a}) =
- \E_{\mu^p}\left[\nabla^2 \log {\mu}(\pmb{s}^\prime|\pmb{s},\pmb{a};{\hat{\pmb{\beta}}_n})\right]$, for Gaussian state transition model is detailed in {Appendix~\ref{appendix: information matrix estimation}}.}


\subsection{Policy Optimization}
\label{sec:policyOptimizationstep}

As explained in Section \ref{sec:framework}, this step produces an updated optimal policy candidate for the real system by using the digital twin $\mathcal{M}^d_n$ constructed in the $n$-th iteration. The process of actually computing this policy is not studied in this paper; we simply assume that, owing to the ease of simulating state transitions for $\mathcal{M}^d_n$, any modern local search RL algorithm such as policy gradient can be applied to learn the policy \citep{ding2024last,cayci2024convergence}. Our focus here is rather on the \textit{design} of $\mathcal{M}^d_n$ to guide policy optimization accounting for model discrepancy.

Based on the digital twin with the updated state transition model $\mu(\pmb{s}^\prime|\pmb{s},\pmb{a};\hat{\pmb{\beta}}_{n})$, the modified MDP, denoted by 
$\Tilde{\mathcal{M}}_{n}^d=(\mathcal{S}, \mathcal{A}, \Tilde{r},{\mu}(\cdot;\hat{\pmb{\beta}}_n),\mu_0)$ in Algorithm~\ref{algo: actor simulator}, differs from   $\mathcal{M}^p=(\mathcal{S}, \mathcal{A}, {r},{\mu}(\cdot;\pmb{\beta}^{\star}),\mu_0)$ of the physical system in two ways. First, the true parameters $\pmb{\beta}^\star$ are replaced by the estimates $\hat{\pmb{\beta}}_n$ computed in the model estimation step (Section~\ref{subsec: calibration setup}). Second, the reward function of the digital twin is augmented as
\begin{equation}\label{eq:augmentedreward}
\tilde{r}\left(\pmb{s},\pmb{a}\right) = r\left(\pmb{s},\pmb{a}\right) - \lambda u\left(\pmb{s},\pmb{a};\hat{\pmb{\beta}}_n,\hat{\pi}_{n-1}\right),
\end{equation}
where $\lambda=c\gamma$ with a tunable parameter $c\in[1,\infty)$ and $u$ is the uncertainty function accounting for the impact of model discrepancy. This function is the key to maintaining the connection between the physical and digital systems. 
It is important to note that $u$ depends on both $\hat{\pmb{\beta}}_n$ and $\hat{\pi}_{n-1}$, that is, both the statistical problem of parameter estimation, and the optimization problem of finding an optimal policy for the physical system, contribute to the uncertainty.

\section{Convergence Analysis} 
\label{subsec: convergence analysis}

The goal of this section is to show the convergence result, i.e., $\lim_{n\rightarrow\infty} J\left(\hat{\pi}_n;\mathcal{M}^p\right) = J\left(\pi^\star;\mathcal{M}^p\right)$, a form of asymptotic optimality, in the \textit{physical} system, of the sequence $\left\{\hat{\pi}_n\right\}$ of policies obtained from the respective digital systems $\left\{\mathcal{M}^d_n\right\}$. 

In our framework, the key link between the physical and digital systems is the uncertainty function $u$. Recall from (\ref{eq:Jbound}) that the uncertainty function bounds the discrepancy between the value of a given policy in the physical and digital systems. The following result presented in Theorem~\ref{thm: uncertainty function convergence} shows that, 
with the proposed Actor-Simulator framework outlined in Algorithm~\ref{algo: actor simulator},
this bound vanishes to zero 
{almost surely (a.s.)} as $n\rightarrow \infty$;
the proof is provided in Appendix~\ref{sec:uncertainty function convergence}.

\begin{theorem}\label{thm: uncertainty function convergence}
Suppose that the sample trajectory $\{(\pmb{s}_i,\pmb{a}_i,\pmb{s}_{i+1})\}_{i\in \mathbb{Z}}$ of the physical system satisfies Assumption~\ref{assumption 4}. Let $\pi$ be a fixed policy, and suppose that $\hat{{\pmb{\beta}}}_n$ is computed according to \eqref{eq: calibration estimator}. Then, for any $(\pmb{s},\pmb{a})\in\mathcal{S}\times\mathcal{A}$, {as $n\rightarrow \infty$}, we have
\begin{equation}
u\left(\pmb{s},\pmb{a};\hat{{\pmb{\beta}}}_n, \pi\right) \xrightarrow{a.s.} 0.
\end{equation}
\end{theorem}

We now connect this result to the asymptotic convergence of the policy $\{\hat{\pi}_n\}$ produced by our framework. Recall that the policy is obtained by optimally solving the MDP $\mathcal{M}^d_n$ whose reward function $\tilde{r}$ is penalized by the uncertainty function as in (\ref{eq:augmentedreward}). Lemma~\ref{lemma: conservative under uncertainty-penalized MDP} shows that this design produces a conservative estimate of the reward earned under the physical system $\mathcal{M}^p$, regardless of the policy.

\begin{lemma}[\citealp{Yu2020mopo}]
\label{lemma: conservative under uncertainty-penalized MDP}
Let Assumption \ref{assumption 4} hold. Fix a policy $\pi$ and an estimate $\hat{\pmb{\beta}}$, and let $\Tilde{\mathcal{M}}$ be an MDP with transition probability model 
$\mu(\cdot\mid\pmb{s},\pmb{a};\hat{\pmb{\beta}})$ and reward function
\begin{equation*}
\tilde{r}\left(\pmb{s},\pmb{a}\right) = r\left(\pmb{s},\pmb{a}\right) - \lambda u\left(\pmb{s},\pmb{a};\hat{\pmb{\beta}},\pi\right).
\end{equation*}
{where $\lambda=c\gamma$ with $c\in[1,\infty)$.} Then, $J(\pi;\Tilde{\mathcal{M}}) \leq J\left(\pi; \mathcal{M}^p\right)$.
\end{lemma}
{
Building on Theorem~\ref{thm: uncertainty function convergence} and Lemma~\ref{lemma: conservative under uncertainty-penalized MDP}, we proceed to establish the asymptotic convergence of policy optimization, as stated in Theorem~\ref{thm: convergence}. To prove this result, we first derive an upper bound for the performance gap, $J(\pi^\star; \mathcal{M}^p)-J(\hat{\pi}_n; \mathcal{M}^p)$, using Lemma~\ref{lemma: conservative under uncertainty-penalized MDP}. Since this bound 
is controlled by the uncertainty function $u$, we demonstrate that it converges almost surely to zero by leveraging Theorem~\ref{thm: uncertainty function convergence}. This approach highlights how uncertainty in the learned parameters diminishes as more data is incorporated, leading to the convergence of the policy performance; see the detailed proof of Theorem~\ref{thm: convergence} in Appendix~\ref{sec:ProofTheorem4}.
}

\begin{theorem}\label{thm: convergence}
{Let Assumption \ref{assumption 4} hold, and suppose that Algorithm~\ref{algo: actor simulator} in Section~\ref{sec:framework} is run for $n$ iterations. Then it holds
    \begin{equation*}
        \lim_{n\rightarrow\infty}J(\hat{\pi}_n; \mathcal{M}^p) =J(\pi^\star; \mathcal{M}^p) \ \text{ {a.s.}}
    \end{equation*}}
\end{theorem}

Lastly, we discuss the rate of convergence for the Actor-Simulator algorithm. The rate is difficult to characterize explicitly because it depends on the statistical error in the calibration step, which was discussed in Section~\ref{subsec: calibration setup}. Nonetheless, if we know the decreasing rate of the calibration error denoted as $\psi(n)$ in Theorem~\ref{thm: rate of convergence}, it is possible to connect it to the convergence rate of $J\left(\hat{\pi}_n;\mathcal{M}^p\right)$ using the following result. The detailed proof of Theorem~\ref{thm: rate of convergence} is provided in Appendix~\ref{appendix sec: proof of rate of convergence}.

\begin{theorem} \label{thm: rate of convergence}
Suppose that we are in the situation of Theorem~\ref{thm: convergence}. Suppose, furthermore, that
\begin{equation*}
E\left[\mathcal{L}\left(\hat{\pmb{\beta}}_n\right) - \mathcal{L}({\pmb{\beta}}^\star)\right] = \mathcal{O}\left(\psi(n)\right).
\end{equation*}
Then, it holds
    \begin{equation*}
        \left|J(\hat{\pi}_n; \mathcal{M}^p) -J(\pi^\star; \mathcal{M}^p)\right| = \mathcal{O}\left(\sqrt{\psi(n)}\right).
    \end{equation*}
\end{theorem}

Under the situation when Assumption~{\ref{Regularity Cond} holds, the rate of the calibration error $\psi(n)=\max\left\{\frac{M_{n,\delta}}{\sqrt{n}},\delta\right\}$ as stated in Theorem \ref{thm:expectedlossrate}.
We refer readers to Table \ref{table: high probability bound review} for some examples of $\psi\left(n\right)$ derived in the literature under various restrictive assumptions. 
In the statistical literature, the law of iterated logarithm has been used to analyze the rate of convergence of maximum likelihood estimation in nonlinear models; for example, \cite{qian2002law} considered logistic regression with independent samples, while a more recent work by \cite{yang2021law} studied generalized linear models under $\rho$-mixing assumptions (stronger than $\alpha$-mixing). However, all of these results are still far more specialized than the general nonlinear dynamics considered in our paper.

\section{Empirical Study}\label{sec: experiments}

In this section, 
we use a biopharmaceutical manufacturing example to evaluate the finite sample performance of the proposed Actor-Simulator algorithm. 
This example was adapted from our previous study \citep{wang2023metabolic} 
on induced pluripotent stem cell (iPSC) culture, which is pivotal crucial in the large-scale manufacturing of cell therapies and regenerative medicines, drug discovery, and tissue engineering \citep{hanna2016advanced,stephanopoulos1998metabolic} due to the fact that iPSC has the ability to differentiate into any cell type in the human body. Our goal in this problem is to control the cell culture process to improve productivity and maximize the expected reward, while ensuring 
cell health and functionality. 
Section~\ref{subsec: kinetic model} describes the physics-based model of cell metabolism that underlies the physical system mechanism 
and describes the MDP formulation. Section~\ref{subsec: results} presents the finite-sample performance results, and Sections~\ref{subsec: exploration}-\ref{subec: optimal policy} provide further discussion and interpretation.

\subsection{iPSC Culture Example}
\label{subsec: kinetic model}


The design and control of iPSC cultures consists of adjusting 
cultivation conditions in a way that enhances cell growth, viability, and production consistency.
The performance of iPSCs is known to be very sensitive to small changes in the environmental conditions 
such as nutrient starvation and too much metabolic waste accumulation leading to reduced yields, compromised cell quality, and potential risks to patients. These challenges cannot be easily overcome with existing cell culture control strategies, which are largely ad hoc in nature \citep{dressel2011effects}.

This problem provides a good example of a situation where practitioners have some structural knowledge about the dynamics of cellular metabolism, but with significant uncertainty remaining about key parameters characterizing the underlying bioprocessing regulation mechanisms. We use the metabolic reaction network model proposed by \cite{wang2023metabolic}, which is composed of differential equations modeling
$30$ reactions of central metabolic pathways, $34$ state dimensions, and $73$ parameters. 
Figure~\ref{fig: culture} gives a schematic diagram of the cell metabolic network considered in the empirical study.
The control variable is a feeding strategy, i.e., the percentage of medium exchange (addition of fresh medium to the bioreactor and removal of spent medium).

\begin{figure}[t]
\begin{center}
\includegraphics[height=3.8in]{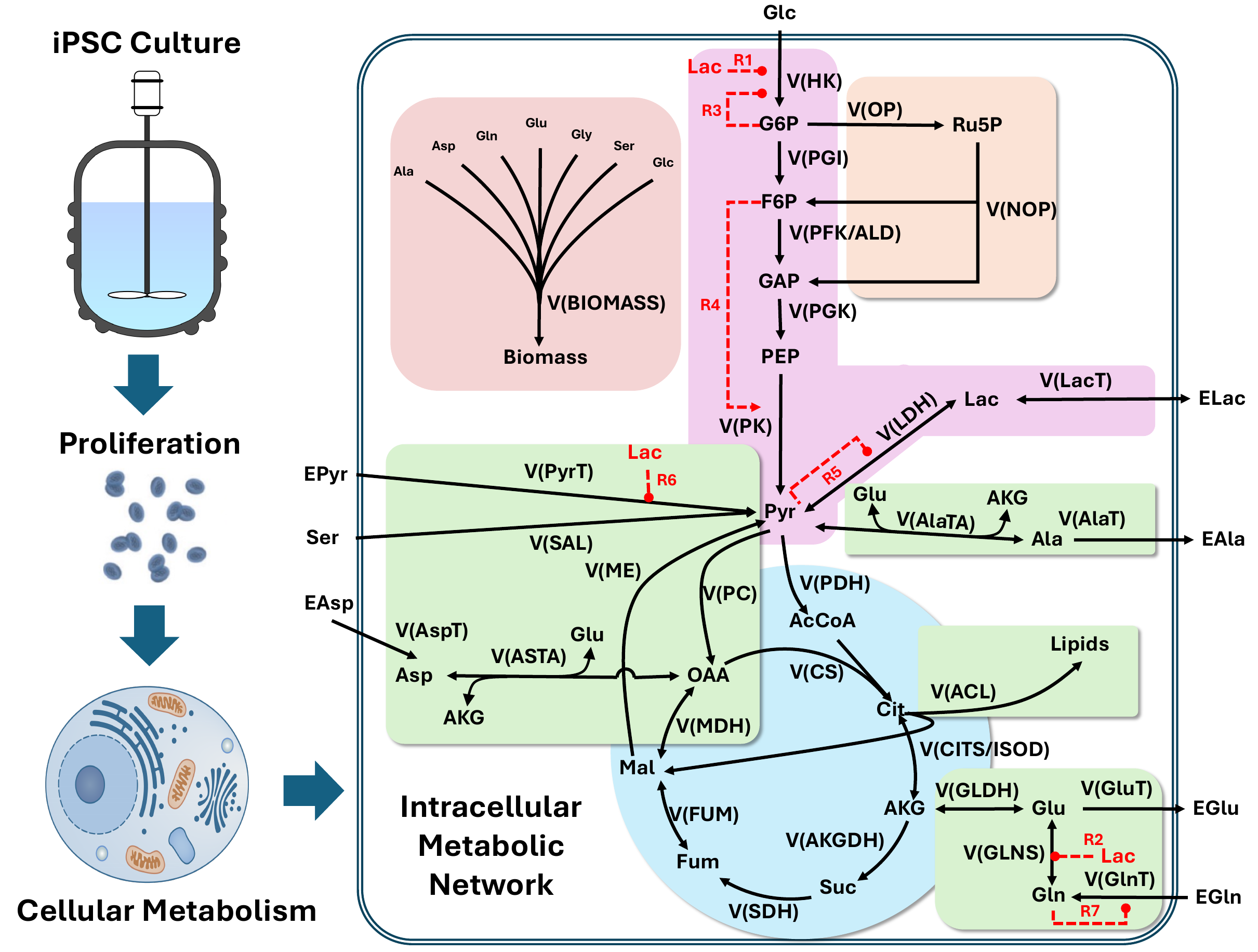}
\caption{Schematic of iPSC culture in bioreactor and medium exchange optimization.}\label{fig: culture}
\end{center}
\end{figure}

We consider dynamic changes in state variables $\pmb{s}(t)=\left(X(t),\pmb{u}(t)^\top\right)^\top$, where $X(t)$ is cell density and $\pmb{u}\left(t\right)$ is a vector representing concentrations of extracellular metabolites at any time $t$. 
Data from the physical system are collected at every 4-hour interval, i.e., $\Delta t=4$ h, that is represented by $(\pmb{s}_n,\pmb{a}_n,\pmb{s}_{n+1})$. In specific, the state at the time of the $n$-th measurement is denoted by $\pmb{s}_n= (X_n, \pmb{u}_n^\top)^\top$. 
The state-transition dynamics of the physical system are expressed by
\begin{equation}
  \pmb{s}_{n+1} = \pmb{s}_n + \Delta t\cdot\pmb{f}(\pmb{s}_n,\pmb{a}_n;\pmb{\pmb{\beta}})+\pmb{\epsilon}, 
  \label{eq.stateTransitionExample}
\end{equation}
where $\pmb{\epsilon}$ represents the inherent stochastic uncertainty of cell culture process induced by uncontrolled factors, following a multivariate Gaussian distribution with zero mean and unknown diagonal covariance $\Sigma$. In the empirical study, the diagonal variance term for each metabolite is set to be 5\% of its initial value. 
The estimation of the conditional Fisher information matrix for this model, $\mathcal{I}(\hat{\pmb\beta}_n|\pmb{s},\pmb{a})$, is detailed in Appendix~\ref{appendix: information matrix estimation}.

In biopharmaceutical manufacturing, the immediate impact of action on the state change is often given. The action depends on the ratio of the spent medium, denoted by a continuous variable $0 \leq b \leq 1$, replaced by the fresh medium. 
The 
feeding decision $\pmb{a}_n=\pi(\pmb{s}_n; b)$ has an immediate 
effect on the change of extracellular cultivation condition 
according to the formula $\pmb{u}_n^+ = b\times 
    \pmb{u}_0 + (1-b)\pmb{u}_n,$
where $\pmb{u}_0$ represents the concentration profile of the fresh medium, $\pmb{u}_n$ and $\pmb{u}_n^+$ represent the extracellular environmental conditions before and after the feeding decision. 
In addition, during the $n$-th time interval, the reward earned by taking action {$\pmb{a}_n$} in state $\pmb{s}_n$ is expressed as
\begin{equation}
r(\pmb{s}_n,\pmb{a}_n) = c_r \Delta X_n
- c_m {b} - c_l \Delta\mbox{[ELAC]}_n,
\label{eq.exampleReward}
\end{equation}
where 
$c_r$ is the revenue per unit of iPSC cell product, 
$c_m$ is the unit cost of the exchange medium, and $c_l$ is a penalty cost for metabolic wastes. The quantities of 
$\Delta X_n$
 and $\Delta\mbox{[ELAC]}_n$
 represent the concentration changes of the target product (i.e., iPSCs) and the metabolic waste (i.e., extracellular lactate)  accumulated during the $\Delta t$ time interval.
 Therefore, our goal is to optimize 
 the objective {$J(\pi;\mathcal{M}^p) =\E[\sum^\infty_{n=0}\gamma^n r(\pmb{s}_n,\pmb{a}_n)]$} 
 balancing yield and production costs
 with $\gamma=0.99$ used in the empirical study.

The digital twin calibration focuses on the mechanistic parameters specifying the underlying bioprocessing mechanisms of iPSC production process.
Therefore, the function $\pmb{f}$ in (\ref{eq.stateTransitionExample}) is built on the existing structural information of cell culture mechanisms \citep{wang2023metabolic}. Specifically, to consider the activation and inhibition effects of the extracellular nutrient (i.e., glucose and glutamine) and the metabolic waste concentration (i.e., lactate) on cell growth, the growth dynamics of the cell density $X(t)$ measured in cells/cm$^2$ is governed by {the following system of differential equations, which provides a simplified mechanistic model from \cite{kornecki2018process},}
\begin{align}
    \frac{\d X(t)}{\d t} &= (\mu - \mu_d) X(t) \label{eq:celldensity} \\ 
    \mu_d &= k_d \cdot \frac{\mbox{[ELAC]}}{\mbox{[ELAC]} + K_{Dlac}} 
    \nonumber \\
    \mu &= \mu_{max} \frac{\mbox{[GLC]}}{K_{glc} + \mbox{[GLC]}} \frac{\mbox{[EGLN]}}{K_{glc} + \mbox{[EGLN]}} \frac{K_{Ilac}}{K_{Ilac} + \mbox{[ELAC]}}
    \nonumber 
\end{align}
where the constants $K_{Dlac}$, $K_{glc}$ and $K_{Ilac}$, the maximum growth rate $\mu_{max}$ and the death rate $k_d$ can be found in Table 2 of \cite{kornecki2018process},
and $\mbox{[A]}$ represents the extracellular concentration of molecule A at time $t$.
In specific, [ELAC] refers to the concentration of extracellular lactate, a byproduct of glycolysis and a key metabolite in energy metabolism. [GLC] denotes the concentration of extracellular glucose, the primary energy source for most cells, and a critical input for glycolysis. [EGLN] represents the concentration of extracellular glutamine, an essential amino acid that serves as a carbon and nitrogen source in various biosynthetic pathways.

In addition to the immediate impact from actions, the dynamics of the concentrations of extracellular metabolites $\pmb{u}\left(t\right)$ depend on cell response to environmental changes, modeled using a vector-valued reaction rate function denoted by $\pmb{v}\left(\pmb{s}\left(t\right); \pmb{\beta}\right)$. Letting $N$ denote a known stoichiometry matrix characterizing the metabolic reaction network structure, we have the system of metabolic  equations, i.e.,
\begin{equation}
    \frac{\d\pmb{u}(t) }{\d t}
    = N  \pmb{v}\left(\pmb{s}(t); \pmb{\beta} \right)  X(t).
    \label{eq:state} \nonumber 
\end{equation}
In the context of iPSC cultures, the metabolic regulation model calibration parameters $\pmb{\beta}$ are embedded inside flux rates $\pmb{v}$, with a full specification given in Appendix~\ref{appendix: mechanistic model}.




\subsection{Experimental Results}\label{subsec: results}

We applied the proposed Actor-Simulator algorithm to three instances of this iPSC culture example presented in Section~\ref{subsec: kinetic model}
with, respectively, $20$, $30$, and $40$ 
calibration parameters. The remaining parameters in each instance were assumed to be known. All parameters (both known and unknown) were set to the values reported in \cite{wang2023metabolic}, summarized in Table \ref{table: calibration parameter}.

\begin{table}[t]
\centering
\caption{Summary of calibration parameters of the iPSC culture simulator model.}\label{table: calibration parameter}
\begin{tabular}{@{}llllll@{}}
\toprule
\multicolumn{6}{c}{20 Parameters in Case Study 1, 2, and 3}                                                                                                                   \\
\midrule
$v_{max,HK}$   & \multicolumn{1}{l|}{2.92} & $v_{max,PGI}$      & \multicolumn{1}{l|}{1.43}  & $v_{max,PFK/ALD}$ &2.16\\ 
$v_{max,PGK}$ & \multicolumn{1}{l|}{4.00} & $v_{max,PK}$   & \multicolumn{1}{l|}{3.98} & $v_{max,fLDH}$ &3.28\\
$v_{max,PyrT}$  & \multicolumn{1}{l|}{0.17} & $v_{max,fLacT}$   & \multicolumn{1}{l|}{2.97}  & $v_{max,OP}$ &0.01\\
$v_{max,NOP}$    & \multicolumn{1}{l|}{0.02}  &$v_{max,PDH}$ & \multicolumn{1}{l|}{0.22} & $v_{max,CS}$ &0.43\\
$v_{max,ME}$       & \multicolumn{1}{l|}{0.51} & $v_{max,fMDH}$  & \multicolumn{1}{l|}{1.44} & $v_{max,GlnT}$  &1.81\\
$K_{m,NH4}$     & \multicolumn{1}{l|}{0.17} & $K_{m,ALA}$      & \multicolumn{1}{l|}{0.20}  & $K_{m,GLC}$  &1.46\\
$K_{m,GLN}$     & \multicolumn{1}{l|}{0.26} & $K_{m,GLU}$     & \multicolumn{1}{l|}{0.30} \\
\midrule
\multicolumn{6}{c}{Additional 10 Parameters in Case Study 2 and 3}                                                     \\ \midrule
$v_{max,fCITS/ISOD}$     & \multicolumn{1}{l|}{1.32} & $v_{max,AKGDH}$      & \multicolumn{1}{l|}{2.84} & $v_{max,SDH}$ &0.32 \\ $v_{max,fFUM}$     & \multicolumn{1}{l|}{0.32} & $v_{max,PC}$   & \multicolumn{1}{l|}{0.06} &  $v_{max,fGLNS}$     &1.14\\
$v_{max,fGLDH}$      & \multicolumn{1}{l|}{0.26}  & $v_{max,fAlaTA}$    & \multicolumn{1}{l|}{0.82} & $v_{max,AlaT}$     &0.47\\
$v_{max,GluT}$    & \multicolumn{1}{l|}{0.17} \\ \midrule
\multicolumn{6}{c}{Additional 10 Parameters in Case Study 3}                                                                                                                                                         \\ \midrule
$K_{m,SER}$     & \multicolumn{1}{l|}{0.01} & $v_{m,SAL}$      & \multicolumn{1}{l|}{0.01} & $K_{m,Ru5P}$  &0.02 \\
$K_{m,PYR}$     & \multicolumn{1}{l|}{0.21} & $K_{m,AcCoA}$    & \multicolumn{1}{l|}{0.09} & $K_{m,OAA}$   &0.08 \\
$K_{m,CIT}$     & \multicolumn{1}{l|}{0.39} & $K_{m,AKG}$      & \multicolumn{1}{l|}{2.92} & $K_{m,MAL}$   &0.11 \\
$K_{m,EGLN}$    & \multicolumn{1}{l|}{1.00} \\ \bottomrule
\end{tabular}
\end{table}


For each unknown parameter, we initially draw the corresponding element of $\hat{\pmb{\beta}}_0$ from a uniform distribution varying from zero to four times its underlying true value. We then train the policy in ``episodes,'' each consisting of a $48$-hour cell culture period in which measurements are taken at $4$-hour intervals. At the beginning of every episode, the state variable is reset to $\pmb{u}_0$, but we retain the data and parameter estimates from all previous episodes. The policy $\hat{\pi}_n$ obtained in the $n$-th episode is evaluated by running it separately inside the physical system for $1000$ episodes to estimate $J(\hat{\pi}_n;\mathcal{M}^p)$. We also recorded the relative error of the calibration parameter estimate, i.e., 
$\Vert \frac{\hat{\pmb{\beta}}_n-\pmb{\beta}^\star}{\pmb{\beta}^\star}\Vert$.

In addition to the Actor-Simulator, we considered two benchmark approaches: (1) a random policy on feeding decision, i.e., $\pmb{a}_n=\pi(\pmb{s}_n; b)$ with $b\sim \mbox{Uniform} (0,1)$;
and (2) 
GP-based Bayesian optimization 
\citep{frazier2018bayesian}. 
With the Bayesian optimization approach, at the beginning of each $n$-th iteration, we fit a Gaussian Process (GP) model using the historical dataset $\mathcal{D}_{n-1}=\{(\pmb{s}_i,\pmb{a}_i,\pmb{s}_{i+1})\}_{i=0}^{n-1}$ with input $(\pmb{s}_{i}, \pmb{a}_{i})$ and the next state prediction mean squared error (MSE) as output,
i.e., $\E_{\pmb{s}^\prime \sim \mu(\cdot|\pmb{s}_i, \pmb{a}_i; \hat{\pmb{\beta}}_{n-1})}[(\pmb{s}^\prime - \pmb{s}_{i+1})^2]$. 
Given the state $\pmb{s}_{n}$ from the previous iteration, we select an action $\pmb{a}_{n}$ to maximize the expected improvement on the model prediction \citep{jones1998efficient}. Then, we observe the next state $\pmb{s}_{n+1}$ from the physical system and update the historical dataset $\mathcal{D}_{n}=\mathcal{D}_{n-1} \cup \{(\pmb{s}_{n}, \pmb{a}_{n}, \pmb{s}_{n+1})\}$. Finally, we use the dataset $\mathcal{D}_{n}$ to update the model parameter estimate $\hat{\pmb{\beta}}_{n}$.

\begin{figure}[ht]
 \centering
 \vspace{-0.1in}
 \subfloat[20 Parameters]{
 \centering
 \includegraphics[width=0.32\textwidth]{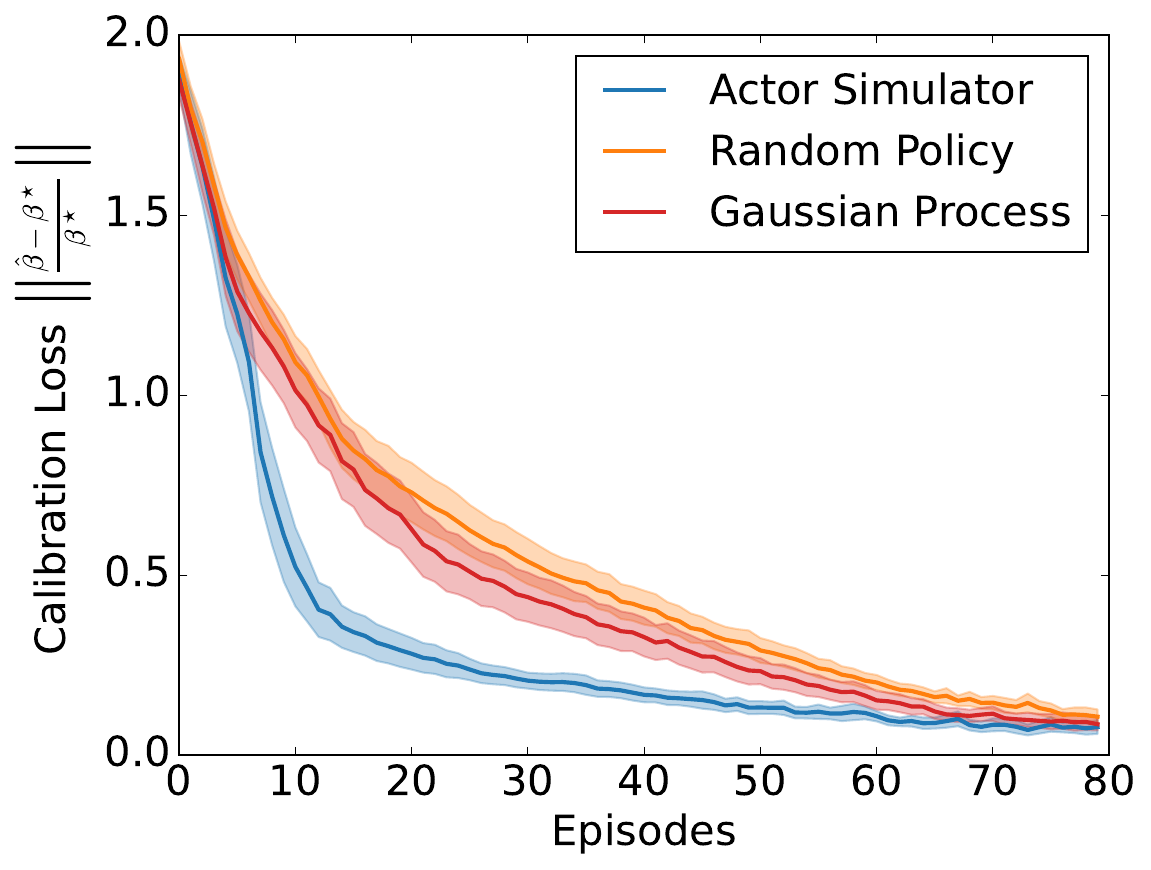}
 }
 \subfloat[30 Parameters]{
 \centering
 \includegraphics[width=0.32\textwidth]{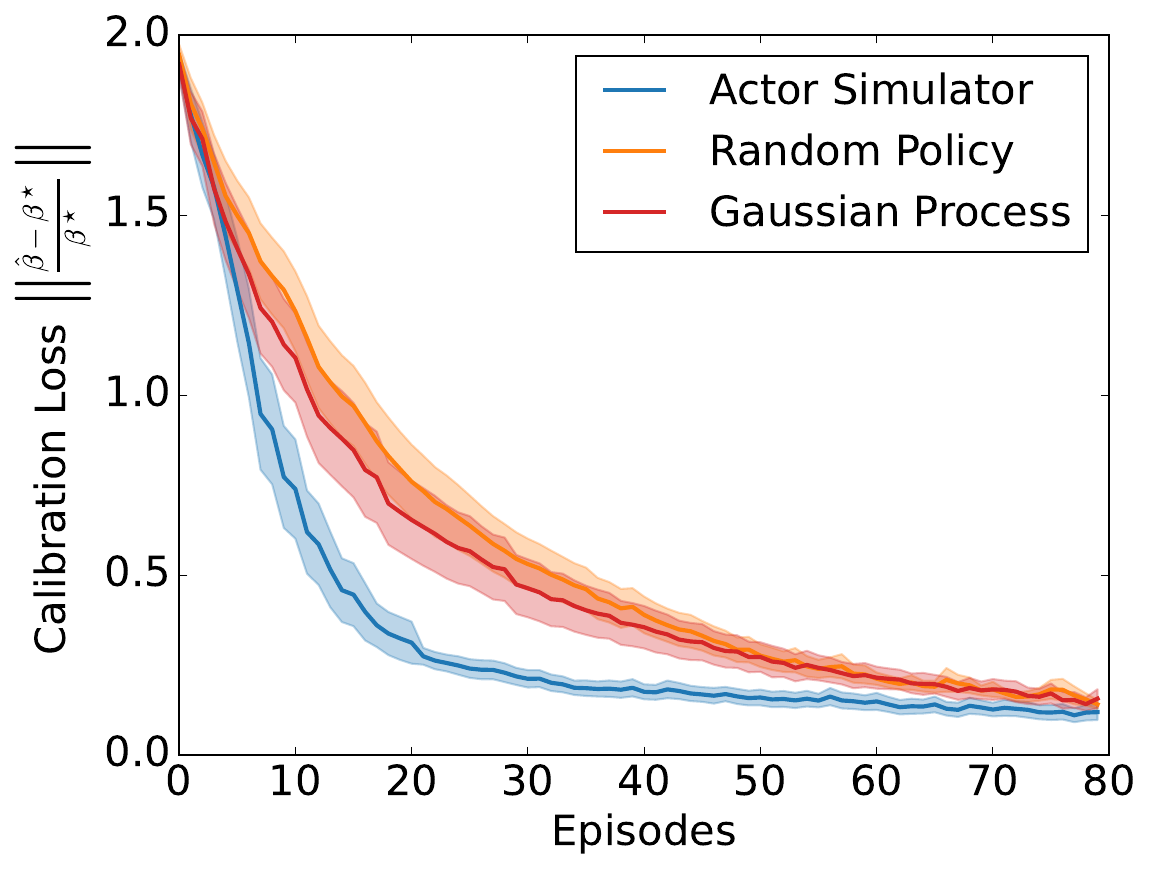}
 }
  \subfloat[40 Parameters]{
 \centering
 \includegraphics[width=0.32\textwidth]{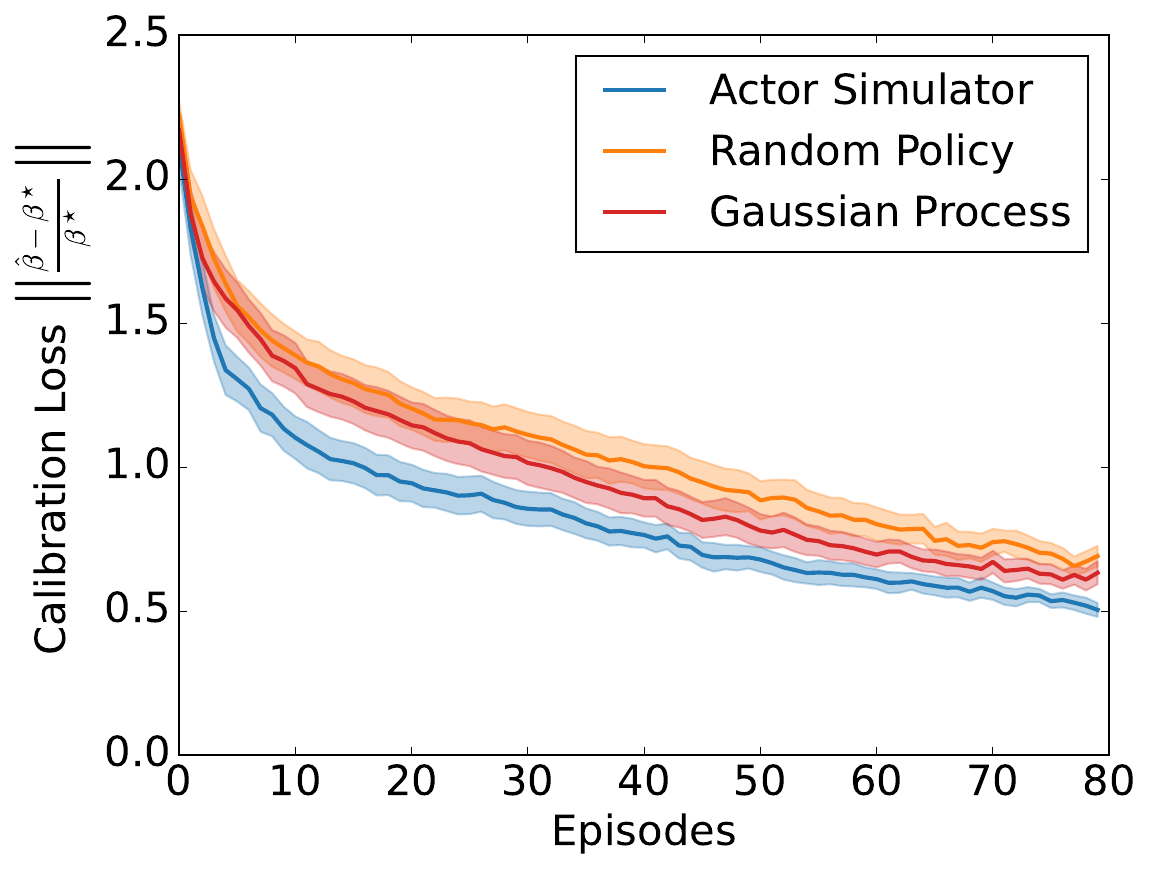}
 }
 \medskip
 \caption{Comparison of calibration parameter estimation performance, i.e.,
 $\Vert \frac{\hat{\pmb{\beta}}_n-\pmb{\beta}^\star}{\pmb{\beta}^\star}\Vert$,
 obtained by three candidate approaches 
 between three calibration settings.
 }\label{fig: calibration performance}
 \vspace{-0.1in}
\end{figure}

\begin{figure}[ht]
 \centering
 \vspace{-0.1in}
 \subfloat[20 Parameters]{
 \centering
 \includegraphics[width=0.32\textwidth]{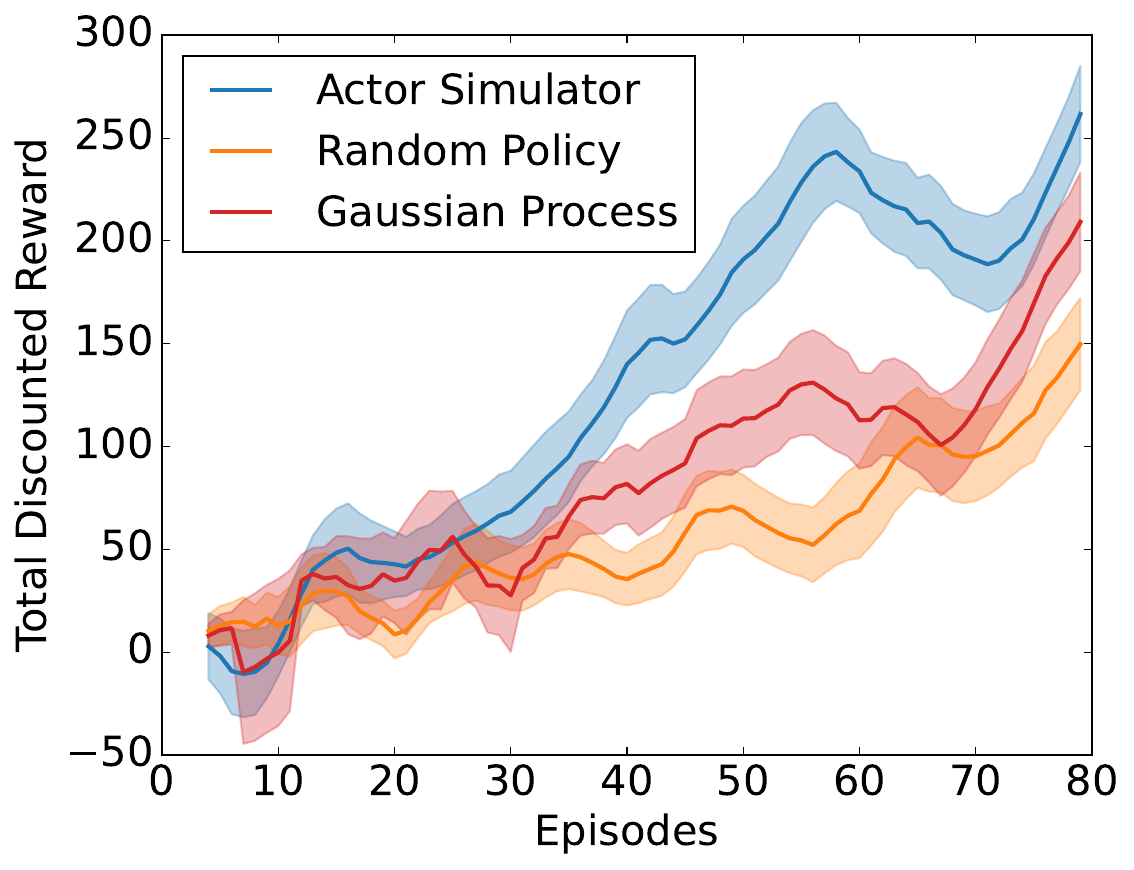}
 }
 \subfloat[30 Parameters]{
 \centering
 \includegraphics[width=0.32\textwidth]{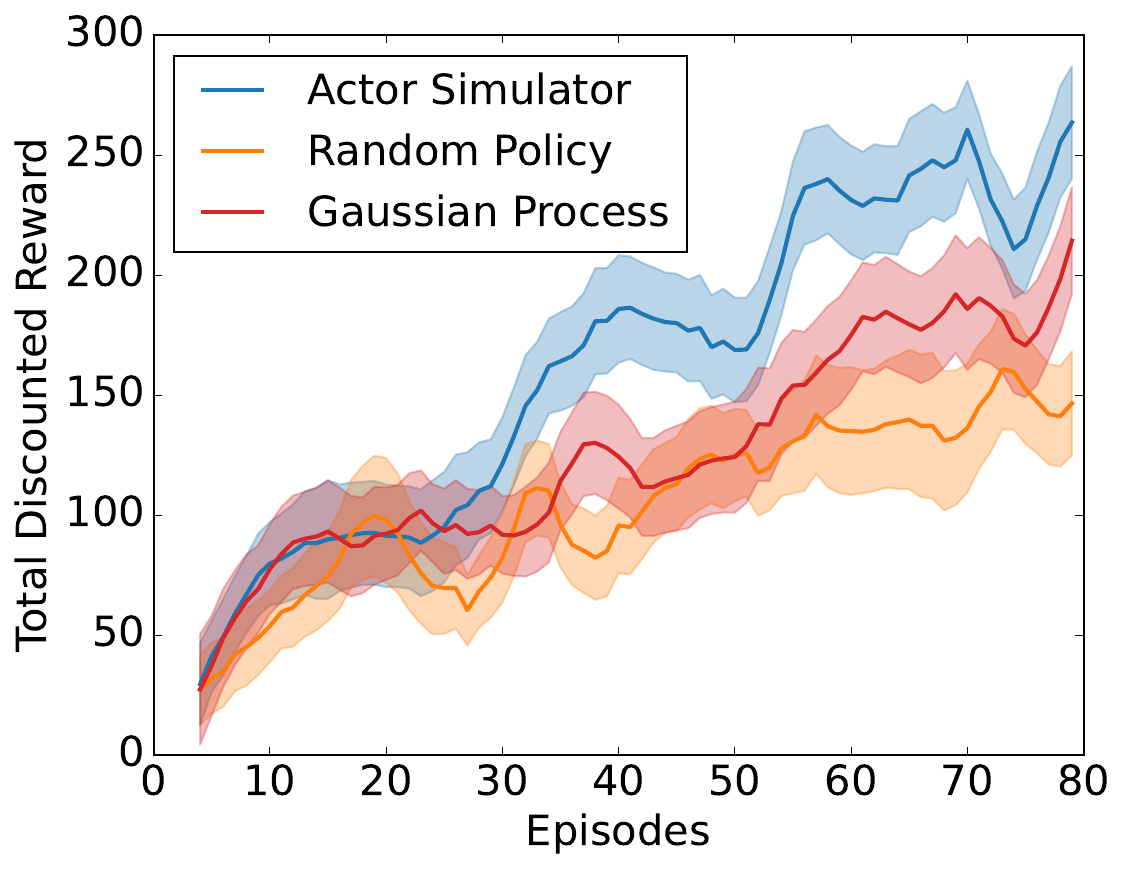}
 }
  \subfloat[40 Parameters]{
 \centering
 \includegraphics[width=0.32\textwidth]{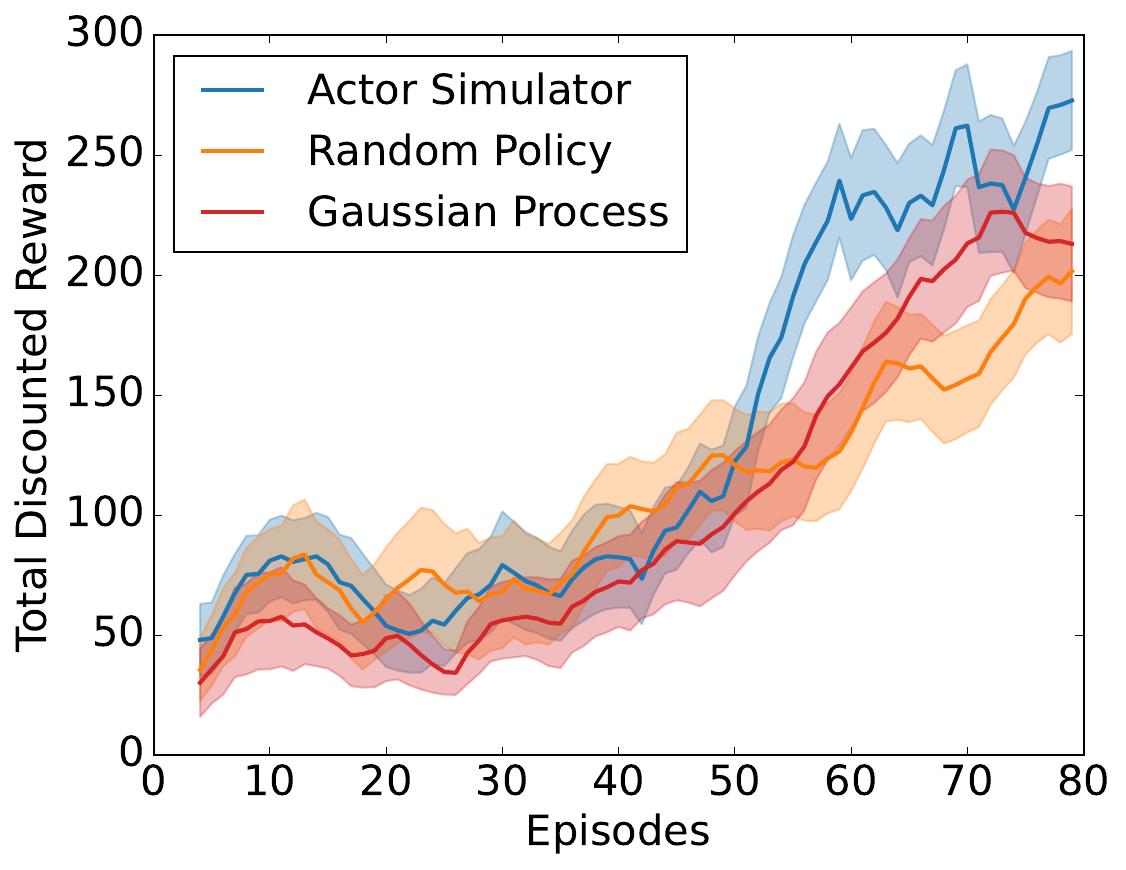}
 }
 \medskip
 \caption{Comparison of policy optimization performance in terms of $J(\hat{\pi}_n; \mathcal{M}^p)$ obtained by three candidate approaches between three calibration settings.
 }\label{fig: RL performance}
 \vspace{-0.1in}
\end{figure}

The performance of all three methods (mean and $95\%$ confidence bounds), with respect to both calibration parameter learning and policy optimization, respectively, is reported in Figures~\ref{fig: calibration performance} and \ref{fig: RL performance}. Each method was run for $30$ macroreplications. 
The results demonstrate that the proposed method consistently outperforms both benchmarks with respect to calibration parameter learning, that translates to significant benefits in policy optimization. 
For the cases with 20 and 30 unknown parameters, the Actor-Simulator reduces the calibration loss to 0.2 using approximately 40.7\% and 48.4\% fewer experiments, respectively, compared to the GP approach. For the case with 40 unknown parameters, the Actor-Simulator consistently outperforms both the random policy and GP by approximately 37.4\% and 37.6\%, respectively, throughout the convergence process.

\subsection{State Space Exploration}\label{subsec: exploration}

\begin{figure}[ht]
 \centering
 \vspace{-0.1in}
 \subfloat[20 Parameters]{
 \centering
 \includegraphics[width=0.33\textwidth]{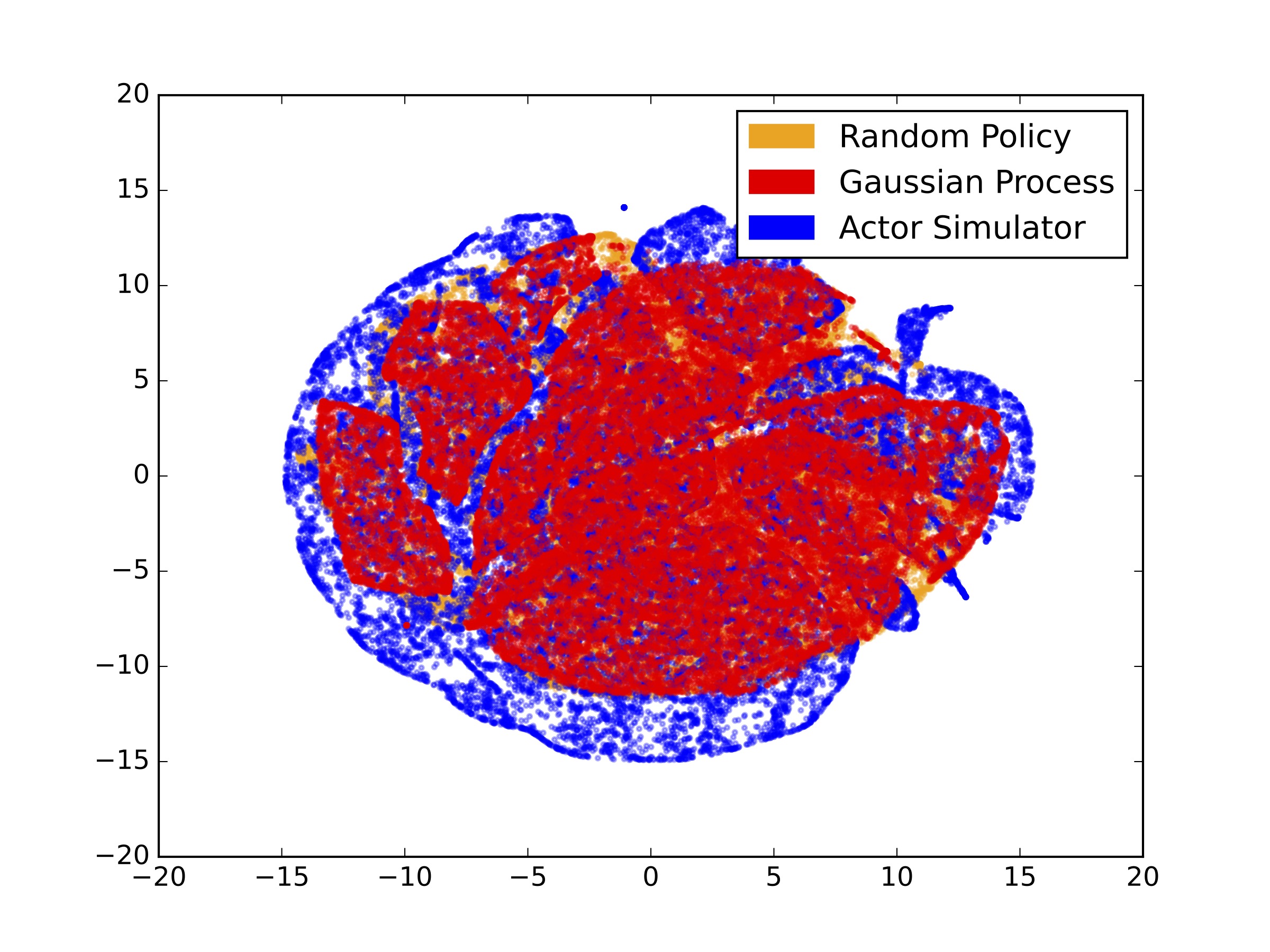}
 }
 \subfloat[30 Parameters]{
 \centering
 \includegraphics[width=0.33\textwidth]{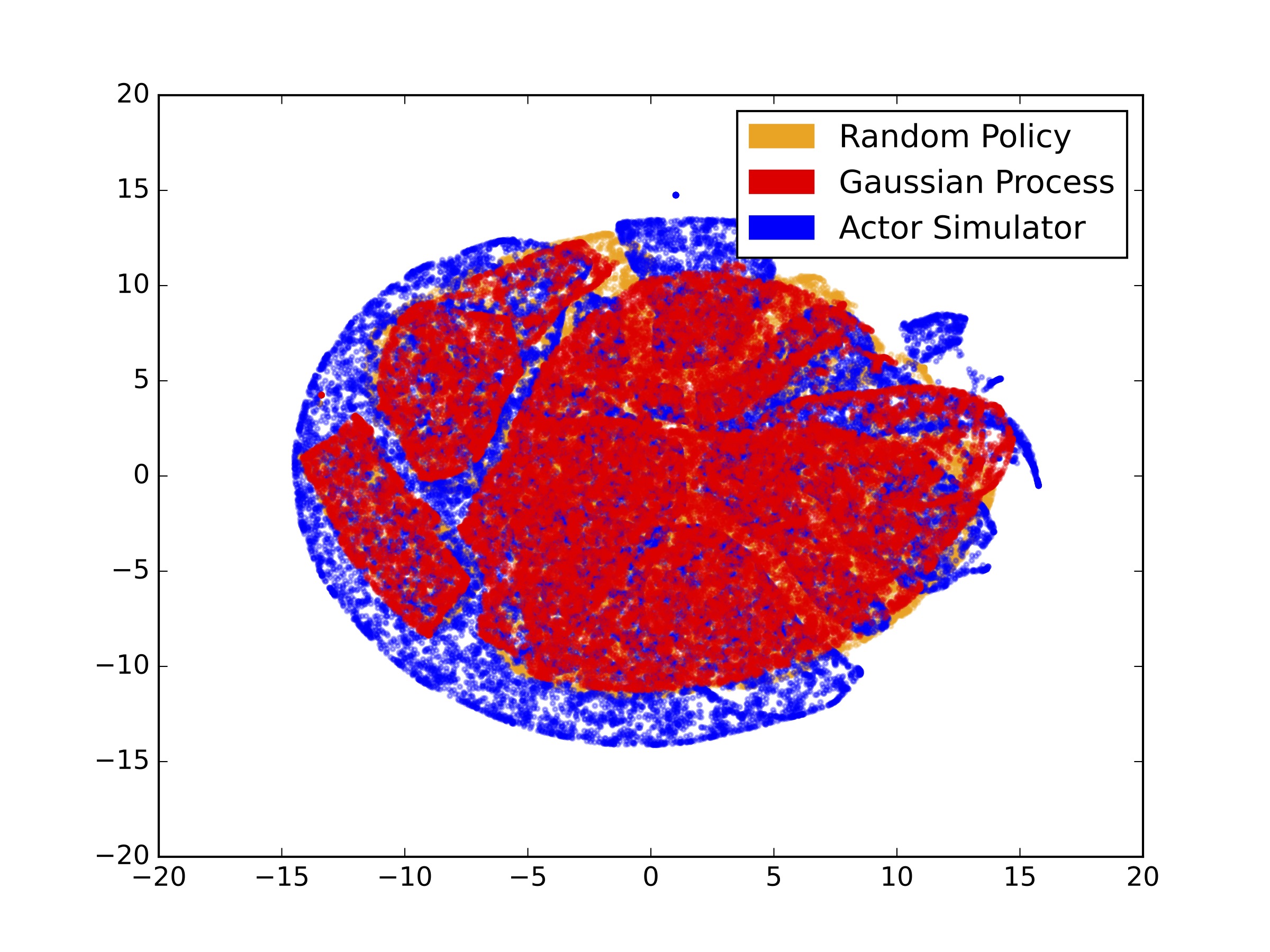}
 }
  \subfloat[40 Parameters]{
 \centering
 \includegraphics[width=0.33\textwidth]{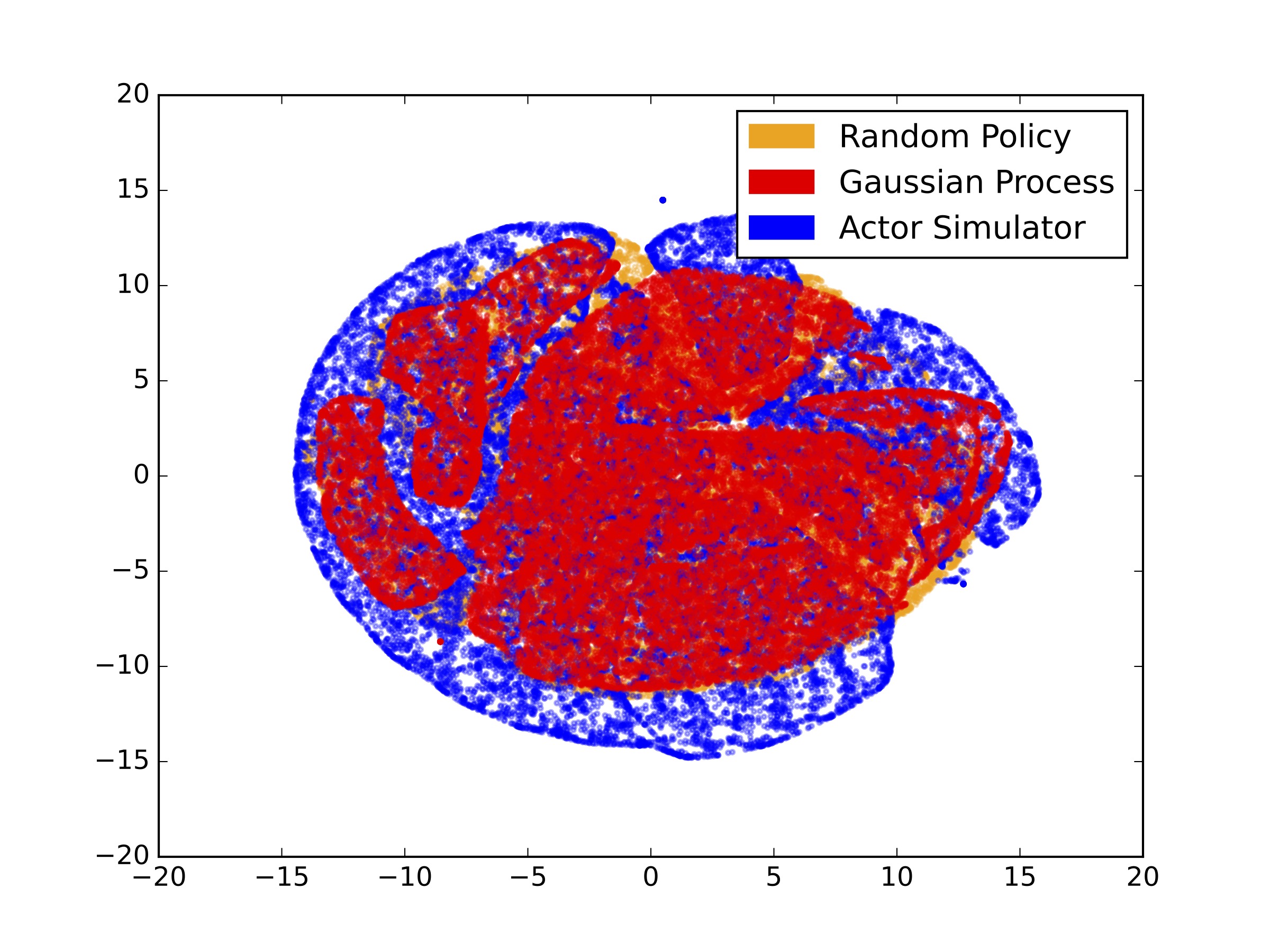}
 }
 \medskip
 \caption{State space exploration of actor-simulator, random policy and Gaussian process based calibration across three calibration settings.
 }\label{fig: tsne-exploration}
 \vspace{-0.1in}
\end{figure}

We used the t-SNE method \citep{van2008visualizing} to project the {$34$-dimensional state space} (see Section~\ref{subsec: kinetic model}) onto the plane, providing a visual representation of the distribution of states explored by each method. These visuals are shown in Figure~\ref{fig: tsne-exploration}. It can be readily seen that the Actor-Simulator algorithm covered a broader range of the state space, consistently reaching further into the outer edges of the plane in each diagram. This also provides insight into the improvements observed in Section~\ref{subsec: results}; one reason why the Actor-Simulator outperforms the benchmarks is because it 
``visits" more of the state space.


\subsection{Prediction Accuracy of Metabolic Pathways}\label{subsec: prediction accuracy of pathways}

Figure \ref{fig: calibration performance of fluxes} visualizes the metabolic network previously shown in Figure \ref{fig: culture}, but labels the metabolic pathways with prediction errors under the proposed method and the GP benchmark. In other words, we compare these two approaches in terms of how well they predict individual flux rates of the cell culture process metabolic network.


\begin{figure}[ht]
 \centering
 \vspace{-0.1in}
 \subfloat[Actor-Simulator]{
 \centering
 \includegraphics[width=0.48\textwidth]{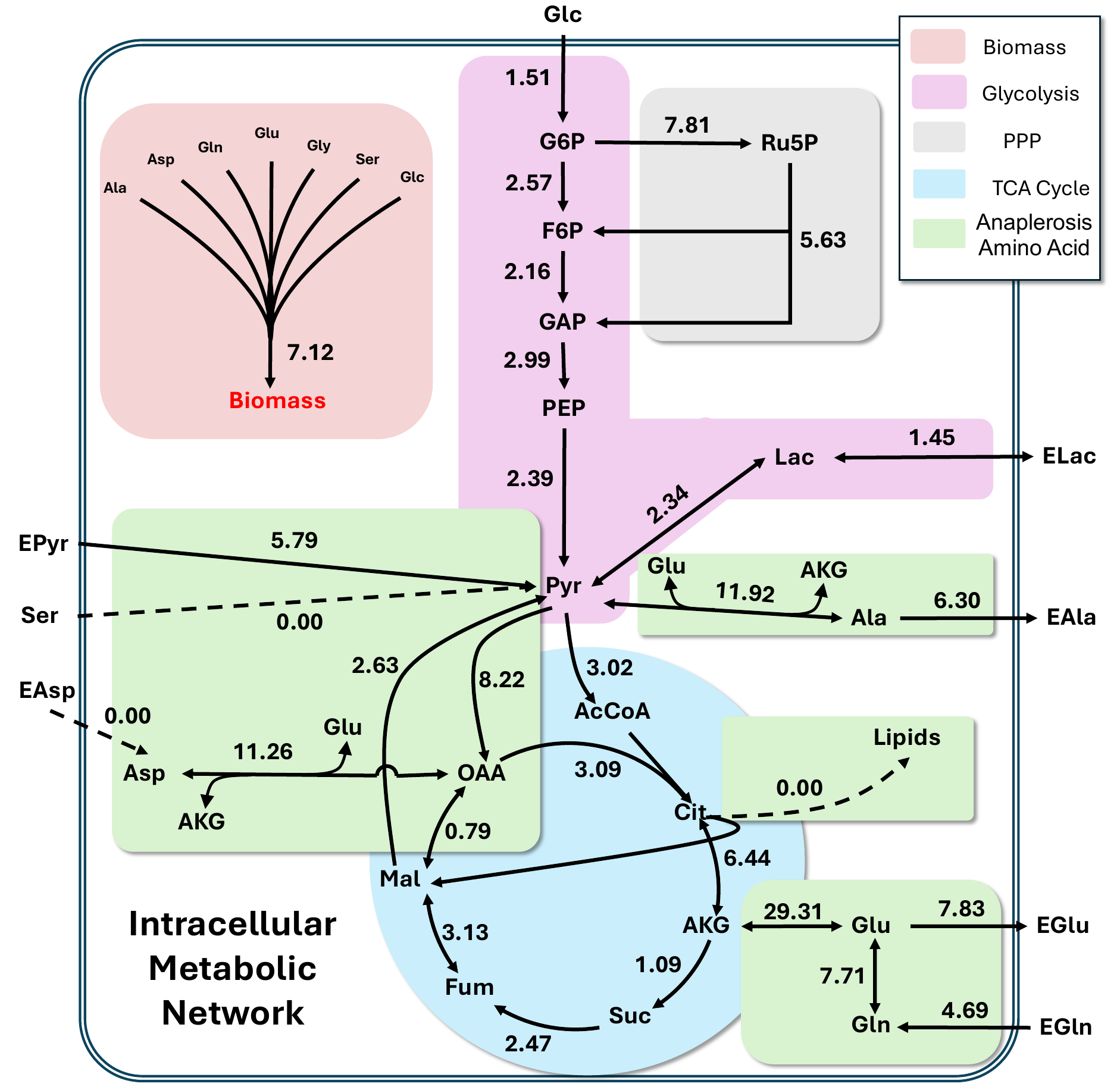}
 }
 \subfloat[Gaussian Process Based Calibration]{
 \centering
 \includegraphics[width=0.48\textwidth]{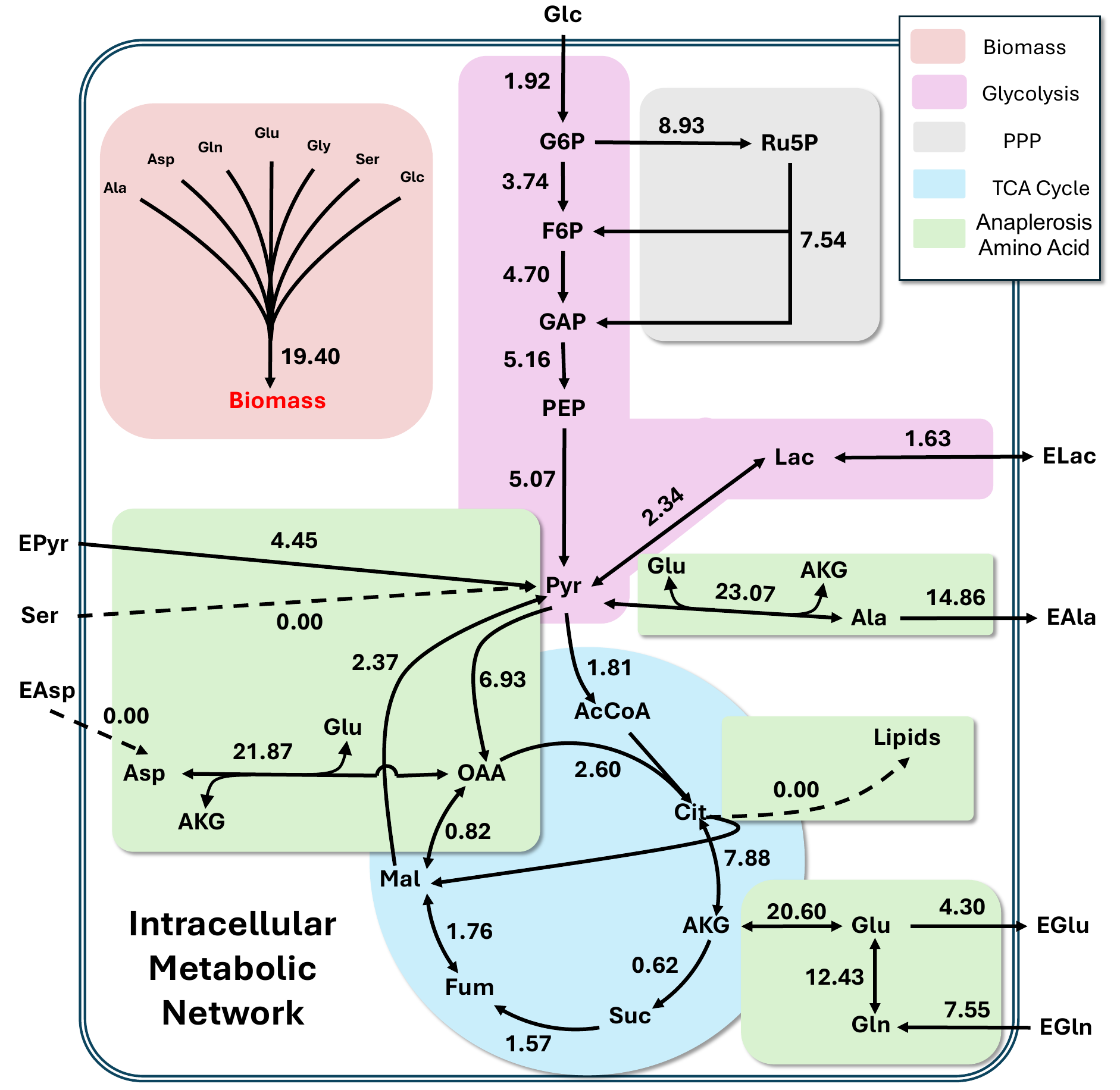}
 }
 \medskip
 \caption{Errors (in percentages) in flux rate predictions made by the Actor-Simulator and GP-based method in the 30-parameter instance, which are measured by mean absolute percentage error (MAPE) between predicted and actual flux rates. The dashed line indicating zero prediction error represents metabolic pathways where the true parameter values are given (not in the set of 30 calibration parameters).
 }\label{fig: calibration performance of fluxes}
 \vspace{-0.1in}
\end{figure}

In general, prediction error is not always lower for the Actor-Simulator. However, the pathways are not all equally important for policy optimization to maximize the objective, i.e., $\arg\max_{\pi\in\Pi} 
J(\pi;\mathcal{M}^p) =\E[\sum^\infty_{n=0}\gamma^n r(\pmb{s}_n,\pmb{a}_n)]$ with the reward defined in (\ref{eq.exampleReward}) and cell growth model (\ref{eq:celldensity}).
For example, the TCA cycle, shown in blue in both diagrams, is generally stable in biomass (i.e.,  
iPSCs) production \citep{zheng2024stochastic} and it is not as crucial to estimate those elements of the parameter vector accurately. By contrast, the biomass synthesis pathway (highlighted in red) is directly related to the primary goal of generating iPSC product. {Glycolysis (highlighted in pink) also plays a crucial role in generating biomass precursors and energy from glucose \citep{rehberg2014glycolysis}. Glutamine, as a key substrate for the tricarboxylic acid (TCA) cycle and amino acid synthesis, supports energy production and the formation of building blocks for cell proliferation. However, excessive glucose and glutamine can lead to the overproduction of metabolic waste (i.e., lactate), 
which can acidify the culture environment and inhibit cell growth. To mitigate these inhibitory effects, precise control of glucose and glutamine concentrations in the medium is essential, balancing nutrient availability to sustain metabolic demand while minimizing lactate generation.}
For these sets of pathways, Actor-Simulator shows substantially lower prediction errors than the GP benchmark. This example illustrates how our calibration approach effectively prioritizes those model parameters that have the highest importance for achieving better feeding policy and RL objective.

\subsection{Medium Exchange Pattern for the Learned Policy} \label{subec: optimal policy}
\begin{figure}[ht]
\begin{center}
\includegraphics[width=1\textwidth]{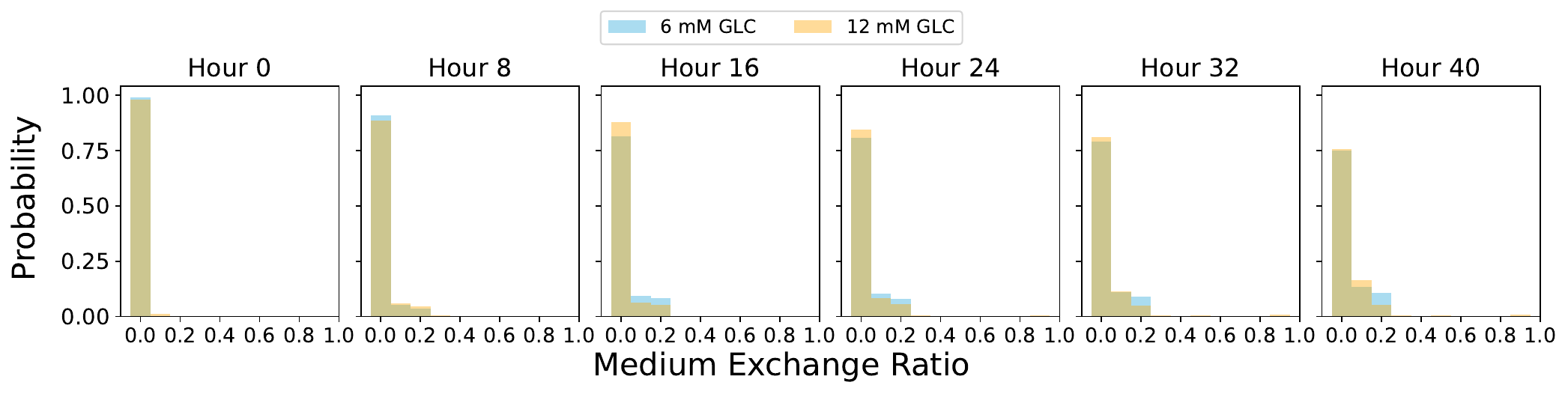}
\caption{The distribution of actions from the optimal medium exchange policy at environment with initial glucose concentration at 6 mM (in blue) and 12 mM (in coral). The values on the x-axis represent the left boundary of an interval of length 0.1; for example, 0.0 represents the interval $[0.0,0.1]$.} \label{fig: distribution at 6mM}
\end{center}
\end{figure}
Here, we demonstrate that the medium exchange policy learned by the Actor-Simulator is interpretable and in line with practitioners' expectations. We examine the policy under two sets of conditions. In both settings, all elements of the state variable are kept constant with the exception of glucose concentration, which can be ``low'' or ``high.'' We take the policy learned for the $30$-parameter problem in Section \ref{subsec: results}, and apply it in the physical system over $12$ time periods. This procedure is repeated $2000$ times across $10$ macroreplications, for a total of $20,000$ experiments. Figure \ref{fig: distribution at 6mM} shows the empirical distribution of the medium exchange ratio chosen by the policy in each of the $12$ time periods.


When the initial glucose concentration is high, we would expect the medium exchange ratio to be lower: the medium already contains adequate nutrition and there is no need to ``feed'' the cells further. Indeed, for the high-glucose setting, we see that the exchange ratios chosen by the policy are mainly clustered around $0$ and $0.1$, with $0.2$ being chosen only occasionally. On the other hand, in the low-glucose setting, the profile is shifted rightward, and nonzero ratios are chosen more frequently. Furthermore, in both settings, the frequency of nonzero ratios is greater in the later time periods, when more of the initial medium has been spent. In short, the actions chosen by the policy are entirely in line with domain knowledge.

\section{Conclusions}\label{sec: conclusion}

This paper introduced the Actor-Simulator, a novel framework for joint digital twin model calibration and policy optimization of complex stochastic systems.
The advantage of a digital twin is that it can easily be run and optimized. However, a policy that is optimal for the digital twin will be suboptimal for the physical system due to the presence of model calibration error. Our approach seeks to collect the most informative data, reduce the calibration error, and accelerate finding an optimal policy for the physical system.
We proved asymptotic optimality of the policy learned by the Actor-Simulator, and empirically demonstrated its effectiveness in finite sample performance 
by using a real-case driven bioharmaceutical manufacturing example.

As the demand for efficient and scalable solutions in biomanufacturing continues to grow, tools like the Actor-Simulator algorithm will play a pivotal role in enhancing production capabilities 
and reducing the costs of 
essential life-saving biopharmaceutical products.
The proposed framework is general and it can be useful for applications in other fields such as 
epidemiological forecasting and control, and large-scale supply chain and manufacturing systems automation, 
where the accuracy and 
cost of digital twin model parameter calibration can significantly influence outcomes.

It is worth emphasizing that the framework considered in this paper is much more general than previous work on calibration \citep{li2019metamodel,treloar2022deep}, which focused on simpler dynamical systems with fewer than 10 parameters and state space dimensions. We also work with general nonlinear dynamics, a setting where parameter estimation poses considerable challenges. In our implementation, we tested multiple fitting methods, including least squares, Nelder-Mead, L-BFGS-B and others, but none of them consistently achieved reliable results. 
Therefore, in the proposed framework, we developed the digital twin in PyTorch and used the Adam optimizer \citep{kingma2015adam} and PyTorch's ``autograd'' feature to compute Hessians and information matrices in digital twin model maximum likelihood estimation.

\begin{appendices}




\section{Proof of Theorem \ref{thm:expectedlossrate}}
\label{exploss}
\textbf{Theorem \ref{thm:expectedlossrate}.} \textit{Suppose that we are in the situation of Lemma \ref{lemma: uniformly tight}. Then,
\begin{equation*}
\E\left[\mathcal{L}\left(\hat{\pmb{\beta}}_n\right) - \mathcal{L}\left(\pmb{\beta}^\star\right)\right] 
= \mathcal{O}\left(\max\left\{\frac{M_{n,\delta}}{\sqrt{n}},\delta\right\}\right).
\end{equation*}}

\subsection{Supporting Lemma}
\textbf{Lemma 2.}\textit{
Let $X_n \defeq \sqrt{n}\left({\mathcal{L}(\hat{\pmb{\beta}}}_n) - \mathcal{L}({\pmb{\beta}}^\star)\right)$. If Assumption \ref{assumption 4} holds, then $\{X_n\}$ is uniformly bounded in probability, i.e., for every $\delta>0$, there exists a large deviation $M_{n,\delta}>0$ such that
\begin{equation*}
\sup_{n\geq 1} P(\Vert X_n\Vert > M_{n,\delta})<\delta, 
\end{equation*}
where $M_{n,\delta}$ depends on both sample size $n$ and tail probability $\delta$.}

\subsection{Proof of the Main Result}
\proof
From Lemma~\ref{lemma: uniformly tight}, we have for every $\delta>0$, there exists a large deviation $M_{n,\delta}>0$ such that the probability of large deviation is less than $\delta$, i.e.,
$\sup_{n\geq 1} P(|X_n| > M_{n,\delta})<\delta$, where $M_{n,\delta}$ depends on both sample size $n$ and tail probability $\delta$.

Thus, there exists an event $\mathcal{E}\defeq 
 \left\{\mathcal{L}(\hat{\pmb{\beta}}_n) - \mathcal{L}({\pmb{\beta}}^\star)\leq \frac{M_{n,\delta}}{\sqrt{n}}\right\}$ such that with $P(\mathcal{E}) \geq 1 - \delta$. Then we can decompose the expectation as
\begin{equation*}
    \E[\mathcal{L}(\hat{\pmb{\beta}}_n) - \mathcal{L}({\pmb{\beta}}^\star)]=\E\left[(\mathcal{L}(\hat{\pmb{\beta}}_n) - \mathcal{L}({\pmb{\beta}}^\star))\mathbf{1}_{\mathcal{E}}\right] + \E\left[(\mathcal{L}(\hat{\pmb{\beta}}_n) - \mathcal{L}({\pmb{\beta}}^\star))\mathbf{1}_{\mathcal{E}^c}\right].
\end{equation*}
For the first term on the event $\mathcal{E}$, we have
\begin{equation*}
    \E\left[(\mathcal{L}(\hat{\pmb{\beta}}_n) - \mathcal{L}({\pmb{\beta}}^\star))\mathbf{1}_{\mathcal{E}}\right]=\E\left[\left.(\mathcal{L}(\hat{\pmb{\beta}}_n) - \mathcal{L}({\pmb{\beta}}^\star))\right|\mathcal{E}\right]= \mathcal{O}\left(\frac{M_{n,\delta}}{\sqrt{n}}\right).
\end{equation*}
{Based on the 
Assumption \ref{assumption 4}(c) presented in Section \ref{subsec: calibration setup}}, 
we know that 
$\mathcal{L}({\pmb{\beta}})=\E\left[\ell(\pmb{s},\pmb{a},\pmb{s}^\prime; {\pmb{\beta}})\right] \leq U_{\ell}$.
Then, for the second term, by H\"{o}lder's inequality, we have
\begin{equation*}
    \E\left[(\mathcal{L}(\hat{\pmb{\beta}}_n) - \mathcal{L}({\pmb{\beta}}^\star))\mathbf{1}_{\mathcal{E}^c}\right]\leq {{\left[\left(\mathcal{L}(\hat{\pmb{\beta}}_n) - \mathcal{L}({\pmb{\beta}}^\star)\right)^2\right]^{1/2}}\delta
    \leq 2U_{\ell} \delta }=\mathcal{O}\left(\delta\right).
\end{equation*}
By putting them together, for $n$
sufficiently large, we obtain the conclusion, 
\begin{equation*}
    \E\left[\mathcal{L}(\hat{\pmb{\beta}}_n) - \mathcal{L}({\pmb{\beta}}^\star)\right] = \mathcal{O}\left(\max\left\{\frac{M_{n,\delta}}{\sqrt{n}},{\delta}\right\}\right).
\end{equation*}
\endproof

\section{Proof of Theorem \ref{thm: uncertainty}}
\textbf{Theorem \ref{thm: uncertainty}.}
\textit{Suppose we are given a policy $\pi \in \Pi$ and an estimate $\hat{\pmb{\beta}}$. Define the uncertainty function of the state-action pair $(\pmb{s},\pmb{a})\in \mathcal{S}\times \mathcal{A}$ as
 \begin{align*}
     &u(\pmb{s},\pmb{a};\hat{{\pmb{\beta}}},\pi)
     \defeq \sqrt{2}\left(1+\log \E_{\mu\left({\cdot}|\pmb{s},\pmb{a};\hat{\pmb{\beta}}\right)}\left[e^{V^\pi(\pmb{s}^\prime;\mathcal{M}^p)^2}\right]\right)^{{1/2}}\sqrt{D_{KL}\left(\mu({\cdot}|\pmb{s},\pmb{a}; {\pmb{\beta}}^\star)\Vert \mu({\cdot}|\pmb{s},\pmb{a}; \hat{\pmb{\beta}})\right)}.
 \end{align*}
Then, we have
\begin{equation*}
   G^\pi(\pmb{s},\pmb{a}) \leq u(\pmb{s},\pmb{a};\hat{{\pmb{\beta}}},\pi)
\end{equation*}
where $G^\pi(\pmb{s},\pmb{a})$ is as in Lemma~\ref{lemma: objective gap between MDPs}.}

\subsection{Supporting Lemmas}

\textbf{Lemma 3} (\cite{Yu2020mopo}, Theorem~4.4).\textit{
Let $\mathcal{M}^p$ and $\mathcal{M}^d$ be two MDPs with the same reward function $r$, but different dynamics $\mu^p$ and $\mu^d$, respectively. For any policy $\pi\in\Pi$, let $G^\pi(\pmb{s},\pmb{a})=\E_{\pmb{s}^\prime\sim\mu^d(\cdot|\pmb{s},\pmb{a})}\left[V^\pi\left(\pmb{s}^\prime;\mathcal{M}^p\right)\right]-\E_{\pmb{s}^\prime\sim{{\mu^p}}(\cdot|\pmb{s},\pmb{a})}\left[V^\pi(\pmb{s}^\prime;\mathcal{M}^p)\right]$. Then,
     $$J\left(\pi; \mathcal{M}^d\right)-J\left(\pi; \mathcal{M}^p\right)=\gamma\mathop{\E}_{(\pmb{s},\pmb{a})\sim d_{\mathcal{M}^d}^{\pi}(s)\pi(\pmb{a}|\pmb{s})}\left[G^\pi(\pmb{s},\pmb{a})\right].$$}

\begin{lemma}[\cite{bolley2005weighted}, Theorem 2.1] \label{lemma: Bolley}
Let $\mathcal{X}$ be a measurable space, let $\mu$, $\nu$ be two
probability measures on $\mathcal{X}$, and let $\phi$ be a nonnegative measurable function on $\mathcal{X}$. Then
\begin{equation*}
    \Vert\phi(\mu-\nu)\Vert \leq \sqrt{2} \left(1+\log\int e^{{\phi(x)}^2} d\nu(x)\right)^{1/2}\sqrt{D_{KL}(\mu\Vert \nu)}
\end{equation*}
where the notation $\phi(\mu-\nu)$ is a shorthand for the signed measure $\phi\mu-\phi\nu$.
\end{lemma}

\subsection{Proof of the Main Result}

\proof
By definition, it holds
    \begin{align}
   G^\pi(\pmb{s},\pmb{a})&=\E_{\pmb{s}^\prime\sim {\mu}(\cdot|\pmb{s},\pmb{a}; \hat{\pmb{\beta}})}\left[V_{\mathcal{M}^p}^\pi(\pmb{s}^\prime)\right]-\E_{\pmb{s}^\prime\sim{\mu}(\cdot|\pmb{s},\pmb{a}; {\pmb{\beta}}^\star)}\left[V_{\mathcal{M}^p}^\pi(\pmb{s}^\prime)\right] \nonumber\\
&=\int_{\mathcal{S}} \left({\mu}( \pmb{s}^\prime|\pmb{s},\pmb{a}; \hat{\pmb{\beta}})- {\mu}(\pmb{s}^\prime|\pmb{s},\pmb{a}; {\pmb{\beta}}^\star)\right) V_{\mathcal{M}^p}^\pi(\pmb{s}^\prime) \d \pmb{s}^\prime\nonumber\\
&\leq \int_{\mathcal{S}} \left |{\mu}(\pmb{s}^\prime| \pmb{s},\pmb{a}; {\pmb{\beta}}^\star)- {\mu}(\pmb{s}^\prime| \pmb{s},\pmb{a}; \hat{\pmb{\beta}})\right| V_{\mathcal{M}^p}^\pi(\pmb{s}^\prime) \d \pmb{s}^\prime \label{eq: weighted TV}
\end{align}
where the likelihoods are represented by 
${\mu}(\pmb{s},\pmb{a}, \pmb{s}^\prime; \hat{\pmb{\beta}})= {\mu}( \pmb{s}^\prime|\pmb{s},\pmb{a}; \hat{\pmb{\beta}})d^{\pi}_{\mathcal{M}^p}(\pmb{s},\pmb{a})$ and ${\mu}(\pmb{s},\pmb{a}, \pmb{s}^\prime; {\pmb{\beta}}^\star)={\mu}( \pmb{s}^\prime|\pmb{s},\pmb{a}; {\pmb{\beta}}^\star)d^{\pi}_{\mathcal{M}^p}(\pmb{s},\pmb{a})$.
{Let $\phi(\pmb{s})\defeq V_{\mathcal{M}^p}^\pi(\pmb{s})$. It is nonnegative because $$\phi(\pmb{s})= V_{\mathcal{M}^p}^\pi(\pmb{s})=\E_{\mu}\left[\left. \sum^\infty_{n=0}\gamma^n r\left(\pmb{s}_n,\pmb{a}_n\right) \right| 
 \pmb{s}_0 = \pmb{s}\right]\geq 0$$ due to the assumption $r(\pmb{s},\pmb{a}) \geq 0$ for any $(\pmb{s},\pmb{a})\in \mathcal{S}\times \mathcal{A}$.}
Then by applying Lemma~\ref{lemma: Bolley}, for any $(\pmb{s},\pmb{a})\in \mathcal{S}\times \mathcal{A}$, \eqref{eq: weighted TV} becomes
\begin{equation*}
   G^\pi(\pmb{s},\pmb{a}) \leq \sqrt{2}\left(1+\log \E_{{\mu}\left(\pmb{s}^\prime|\pmb{s},\pmb{a};\hat{\pmb{\beta}}\right)}\left[e^{V_{\mathcal{M}^p}^\pi(\pmb{s}^\prime)^2}\right]\right)^{{1/2}}\sqrt{D_{KL}\left({\mu}({\cdot}|\pmb{s},\pmb{a}; {\pmb{\beta}}^\star\Vert {\mu}({\cdot}|\pmb{s},\pmb{a}; \hat{\pmb{\beta}})\right)}.
\end{equation*}
\endproof

\section{Proof of Theorem \ref{thm: uncertainty function convergence}}
\label{sec:uncertainty function convergence}

\textbf{Theorem  \ref{thm: uncertainty function convergence}.} \textit{Suppose that the sample trajectory $\{(\pmb{s}_i,\pmb{a}_i,\pmb{s}_{i+1})\}_{i\in \mathbb{Z}}$ of the physical system satisfies Assumption~\ref{assumption 4}. Let $\pi$ be a fixed policy, and suppose that $\hat{{\pmb{\beta}}}_n$ is computed according to \eqref{eq: calibration estimator}. Then, for any $(\pmb{s},\pmb{a})\in\mathcal{S}\times\mathcal{A}$, we have
\begin{equation*}
u\left(\pmb{s},\pmb{a};\hat{{\pmb{\beta}}}_n, \pi\right) \rightarrow 0
\end{equation*}
almost surely.}

\subsection{Supporting Lemma}
\label{subsec:Supportive Lemma}

\textbf{Lemma 1}(\cite{tinkl2013asymptotic}, Corollary 4.3.12). \textit{Let the trajectory $\{(\pmb{s}_i,\pmb{a}_i,\pmb{s}_{i+1})\}_{i\in \mathbb{Z}}$ satisfies Assumption~\ref{assumption 4}(a-d). Then for the estimator $\hat{{\pmb{\beta}}}_n$ defined
 by \eqref{eq: calibration estimator}, we have
 $$\hat{{\pmb{\beta}}}_n\xrightarrow{a.s.} {\pmb{\beta}}^\star.$$
In addition, if Assumption~\ref{assumption 4}(e-h) are met, we further have
     $$\sqrt{n}(\hat{{\pmb{\beta}}}_n - {\pmb{\beta}}^\star) \xrightarrow{d} \mathcal{N}(0,\Sigma({\pmb{\beta}}^\star))$$
where $\Sigma({\pmb{\beta}}^\star) = -\E[\nabla^2 \ell(\pmb{s},\pmb{a},\pmb{s}^\prime;{\pmb{\beta}}^\star)^\top]^{-1}$ is the inverse of Fisher information matrix.}

\subsection{Proof of Theorem~\ref{thm: uncertainty function convergence}}
\begin{proof} 
Under the settings from 
Section \ref{subsec: setting}, i.e., $\sup_{\pmb{s}\in\mathcal{S},\pmb{a}\in\mathcal{A}} r\left(\pmb{s},\pmb{a}\right) \leq r^{\max}$ for some $r^{\max}<\infty$, the value function is bounded, i.e.,
\begin{equation*}
 V^{\pi}_{\mathcal{M}}(s) = \E\left[\sum_{t=0}^\infty\gamma^t r(\pmb{s}_t,\pmb{a}_t) | \pmb{s}_0\right] \leq \sum^\infty_{t=0} \gamma^t r_{max} = \frac{r_{max}}{1-\gamma}.
\end{equation*}
By the definition of the uncertainty function, we have
\begin{align}
    &u(\pmb{s},\pmb{a};\hat{{\pmb{\beta}}}_n, \pi) \nonumber\\
    &= \sqrt{2}\left(1+\log E_{{\mu}\left(\pmb{s}^\prime|\pmb{s},\pmb{a};\hat{\pmb{\beta}}_n\right)}\left[e^{V_{\mathcal{M}^p}^\pi(\pmb{s}^\prime)^2}\right]\right)^{{1/2}}\sqrt{D_{KL}\left({\mu}({\cdot}|\pmb{s},\pmb{a}; {\pmb{\beta}}^\star)\Vert {\mu}({\cdot}|\pmb{s},\pmb{a}; \hat{\pmb{\beta}}_n)\right)} \nonumber \\
    &\leq \sqrt{2} \left(1+\frac{r_{max}}{1-\gamma}\right)^{{1/2}}\sqrt{D_{KL}\left({\mu}({\cdot}|\pmb{s},\pmb{a}; {\pmb{\beta}}^\star)\Vert {\mu}({\cdot}|\pmb{s},\pmb{a}; \hat{\pmb{\beta}}_n)\right)}.
    \label{appendix eq: KL divergence} 
\end{align}
Under the regularity conditions presented in Assumption~\ref{assumption 4},
by applying Lemma \ref{lemma: asymptotic convergence} and the continuous mapping theorem, we have for any tuple $(\pmb{s},\pmb{a},\pmb{s}^\prime)\in\mathcal{S}\times\mathcal{A}\times \mathcal{S}$
 \begin{equation*}
{{\mu}(\pmb{s}^\prime| \pmb{s},\pmb{a}; \hat{{\pmb{\beta}}}_n) \xrightarrow{a.s.} {\mu}(\pmb{s}^\prime| \pmb{s},\pmb{a}; {{\pmb{\beta}}}^\star) \text{ as } n\rightarrow \infty}.
 \end{equation*}
{Based on the settings from Section \ref{subsec: setting}}, the likelihood function ${\mu}(\pmb{s}^\prime| \pmb{s},\pmb{a}; \hat{{\pmb{\beta}}}_n)$ is uniformly bounded. In addition, Assumption~\ref{assumption 4}(b) also implies the log-likelihood function $\log {\mu}(\pmb{s}^\prime|\pmb{s},\pmb{a}; {\pmb{\beta}})$ is also bounded for any ${\pmb{\beta}}\in\mathbb{B}$ and $(\pmb{s},\pmb{a},\pmb{s}^\prime)\in\mathcal{S}\times\mathcal{A}\times\mathcal{S}$. Then applying the bounded convergence theorem leads to
\begin{align}
 &\lim_{n\rightarrow \infty} D_{KL}\left({\mu}({\cdot}|\pmb{s},\pmb{a}; {\pmb{\beta}}^\star)\Vert {\mu}({\cdot}|\pmb{s},\pmb{a}; \hat{\pmb{\beta}}_n)\right) \nonumber\\
 &= \lim_{n\rightarrow \infty} \int_{\mathcal{S}\times\mathcal{A}\times\mathcal{S}}{\mu}(\pmb{s}^\prime|\pmb{s},\pmb{a}; {\pmb{\beta}}^\star)\left[\log {\mu}(\pmb{s}^\prime|\pmb{s},\pmb{a}; {\pmb{\beta}}^\star) - \log {\mu}(\pmb{s}^\prime|\pmb{s},\pmb{a}; \hat{\pmb{\beta}}_n)\right]\d \pmb{s}\d \pmb{a}\d \pmb{s} \nonumber\\
&=\int_{\mathcal{S}\times\mathcal{A}\times\mathcal{S}}{\mu}(\pmb{s}^\prime|\pmb{s},\pmb{a}; {\pmb{\beta}}^\star)\left[\log {\mu}(\pmb{s}^\prime|\pmb{s},\pmb{a}; {\pmb{\beta}}^\star) - \lim_{n\rightarrow \infty} \log {\mu}(\pmb{s}^\prime|\pmb{s},\pmb{a}; \hat{\pmb{\beta}}_n)\right]\d \pmb{s}\d \pmb{a}\d \pmb{s} \nonumber\\
&=0 \ \text{ a.s.} \label{appendix eq: KL divergence convergence}
\end{align}
Therefore, the conclusion $u\left(\pmb{s},\pmb{a};\hat{{\pmb{\beta}}}_n, \pi\right) \xrightarrow{a.s.} 0$ as $n\rightarrow \infty$ follows by applying \eqref{appendix eq: KL divergence convergence} to the upper bound \eqref{appendix eq: KL divergence}.
\end{proof}

\section{Proof of Theorem~\ref{thm: convergence}}
\label{sec:ProofTheorem4}

\textbf{Theorem~\ref{thm: convergence}.} 
\textit{Let Assumption \ref{assumption 4} hold, and suppose that Algorithm~\ref{algo: actor simulator} in Section~\ref{sec:framework} is run for $n$ iterations. Then it holds
    \begin{equation*}
        \lim_{n\rightarrow\infty}J(\hat{\pi}_n; \mathcal{M}^p) =J(\pi^\star; \mathcal{M}^p) \ \text{ {a.s.}}
    \end{equation*}}

\subsection{Supporting Lemma}

\textbf{Lemma 4} (\cite{Yu2020mopo}).\textit{
Let Assumption \ref{assumption 4} hold. Fix a policy $\pi$ and an estimate $\hat{\pmb{\beta}}$, and let $\Tilde{\mathcal{M}}$ be an MDP with transition probability model 
$\mu\left(\cdot\mid\pmb{s},\pmb{a};\hat{\pmb{\beta}}\right)$ and reward function
\begin{equation*}
\tilde{r}\left(\pmb{s},\pmb{a}\right) = r\left(\pmb{s},\pmb{a}\right) - \lambda u\left(\pmb{s},\pmb{a};\hat{\pmb{\beta}},\pi\right).
\end{equation*}
{where $\lambda=c\gamma$ with $c\in[1,\infty)$.} Then, $J(\pi;\Tilde{\mathcal{M}}) \leq J\left(\pi; \mathcal{M}^p\right)$.}

\proof
The conclusion holds because
    \begin{align}
    J\left(\pi; \mathcal{M}^p\right) &
   {=} J\left(\pi; \mathcal{M}^d\right)- \gamma \mathop{\E}_{(\pmb{s},\pmb{a})\sim d_{\mathcal{M}^d}^{\pi}(\pmb{s})\pi(\pmb{a}|\pmb{s})}\left[G^\pi(\pmb{s},\pmb{a})\right]\label{eq: lemma 4 equation 1}\\
    &=   \mathop{\E}_{(\pmb{s},\pmb{a})\sim d_{\mathcal{M}^d}^{\pi}(\pmb{s})\pi(\pmb{a}|\pmb{s})}\left[r(\pmb{s},\pmb{a})-\gamma G^\pi(\pmb{s},\pmb{a})\right] \nonumber\\
    &\geq \mathop{\E}_{(\pmb{s},\pmb{a})\sim d_{\mathcal{M}^d}^{\pi}(s)\pi(\pmb{a}|\pmb{s})}\left[r(\pmb{s},\pmb{a})-\gamma u(\pmb{s},\pmb{a};\hat{{\pmb{\beta}}}, 
    \pi)\right] \label{eq:uncertainty reward}\\
    &{\geq} \mathop{\E}_{(\pmb{s},\pmb{a})\sim d_{\mathcal{M}^d}^{\pi}(s)\pi(\pmb{a}|\pmb{s})}\left[\Tilde{r}(\pmb{s},\pmb{a})\right] \label{eq:uncertainty reward constant}\\
   & \eqdef J(\pi;\Tilde{\mathcal{M}})\nonumber
\end{align}
where \eqref{eq: lemma 4 equation 1} holds due to Lemma~\ref{lemma: objective gap between MDPs} and \eqref{eq:uncertainty reward} holds due to Theorem~\ref{thm: uncertainty}. {Finally, \eqref{eq:uncertainty reward constant} holds because $\lambda = \gamma c$ with $c \in [1,\infty)$.}
\endproof

\subsection{Proof of Theorem~\ref{thm: convergence}} 
\proof
Let $\epsilon_u(\pi)\defeq {\E_{(\pmb{s},\pmb{a})\sim d_{\mathcal{M}^d}^{\pi}(\pmb{s})\pi(\pmb{a}|\pmb{s})}}[u(\pmb{s},\pmb{a}; \hat{{\pmb{\beta}}}_n, \pi)]$. From \eqref{eq:Jbound}, it holds
 \begin{equation}
     J(\pi;\mathcal{M}^d)  \geq J(\pi;\mathcal{M}^p)-{\gamma}\epsilon_u(\pi). \label{eq: eq 0 thm 2}
 \end{equation}
 Then for any policy $\pi$, we have, 
 \begin{align}
    J(\hat{\pi}_n;\mathcal{M}^p) &\geq J(\hat{\pi}_n;\Tilde{\mathcal{M}}) \label{eq: eq 1 thm 2}\\
    &\geq J(\pi;\Tilde{\mathcal{M}}) \label{eq: eq 2 thm 2} \\
    &
    =\mathop{\E}_{(\pmb{s},\pmb{a})\sim d_{\mathcal{M}^d}^{\pi}(s)\pi(\pmb{a}|\pmb{s})}\left[r(\pmb{s},\pmb{a})-\lambda u(\pmb{s},\pmb{a};\hat{{\pmb{\beta}}}_n, 
    \pi)\right]\nonumber\\
    &{= J\left(\pi;\mathcal{M}^d\right) - \lambda \epsilon_u(\pi)}
    \nonumber\\
    &\geq J(\pi;\mathcal{M}^p) - (\lambda+\gamma) \epsilon_u(\pi)\label{eq: eq 4 thm 2}
 \end{align}
 where \eqref{eq: eq 1 thm 2} follows Lemma~\ref{lemma: conservative under uncertainty-penalized MDP}, \eqref{eq: eq 2 thm 2} 
 holds by 
 the definition of the optimal policy $\hat{\pi}$ w.r.t. the uncertainty-penalty MDP, and \eqref{eq: eq 4 thm 2} follows by applying {
 \eqref{eq: eq 0 thm 2}.}
 

 Now let's take a look at the almost sure convergence of the policy optimization. Let $\pi^\star = \argmax_{\pi\in\Pi}J(\pi; \mathcal{M}^p)$ be the optimal policy. Since \eqref{eq: eq 4 thm 2} holds for any policy $\pi$, 
 we have 
 \begin{align}
 {J(\hat{\pi}_n; \mathcal{M}^p) \geq J(\pi^\star; \mathcal{M}^p) - (\lambda+\gamma)\epsilon_u(\pi^\star)}\label{eq: theo 4 conv}.
 \end{align}
From Theorem~\ref{thm: uncertainty function convergence} which shows that $\lim_{n\rightarrow \infty} {u(\pmb{s},\pmb{a};\hat{\pmb{\beta}}_n,\pi)}=0$ a.s. for any policy $\pi\in \Pi$ and $(\pmb{s},\pmb{a})\in\mathcal{S}\times \mathcal{A}$, we have
\begin{eqnarray*}
  \lim_{n\rightarrow \infty}\epsilon_u(\pi)&=& \lim_{n\rightarrow \infty}\int_{\mathcal{S}}\int_{\mathcal{A}}u(\pmb{s},\pmb{a};\hat{\pmb{\beta}}_n, \pi) 
  d_{\mathcal{M}^d}^{\pi}(\pmb{s})\pi(\pmb{a}|\pmb{s})
  \d \pmb{s} \d \pmb{a}  \\
  &\leq& \lim_{n\rightarrow \infty}\int_{\mathcal{S}}\int_{\mathcal{A}}u(\pmb{s},\pmb{a};\hat{\pmb{\beta}}_n, \pi)  \d \pmb{s} \d \pmb{a}.
\end{eqnarray*}
Note that the sequence of functions $u(\pmb{s},\pmb{a};\hat{\pmb{\beta}}_n, \pi)$ is uniformly bounded on $\mathcal{S}\times\mathcal{A}$. The conclusion is obtained by applying the bounded convergence theorem
\begin{equation*}
  \lim_{n\rightarrow \infty}\epsilon_u(\pi) \leq \lim_{n\rightarrow \infty}\int_{\mathcal{S}}\int_{\mathcal{A}}u(\pmb{s},\pmb{a};\hat{\pmb{\beta}}_n, \pi)  \d \pmb{s} \d \pmb{a} = \int_{\mathcal{S}}\int_{\mathcal{A}} \lim_{n\rightarrow \infty} u(\pmb{s},\pmb{a};\hat{\pmb{\beta}}_n, \pi)  \d \pmb{s} \d \pmb{a}=0.
\end{equation*}
\endproof
\section{Proof of Theorem~\ref{thm: rate of convergence}}
\label{appendix sec: proof of rate of convergence}

\textbf{Theorem~\ref{thm: rate of convergence}.} 
\textit{Under the same condition as Theorem~\ref{thm: convergence}, suppose, furthermore, that
\begin{equation}
E\left[\mathcal{L}\left(\hat{\pmb{\beta}}_n\right) - \mathcal{L}({\pmb{\beta}}^\star)\right] = \mathcal{O}\left(\psi(n)\right).
\label{eq.calibrationLoss}
\end{equation}
Then,
    \begin{equation*}
        \left|J(\hat{\pi}_n; \mathcal{M}^p) -J(\pi^\star; \mathcal{M}^p)\right| = \mathcal{O}\left(\sqrt{\psi(n)}\right).
    \end{equation*}}

\noindent\proof
From \eqref{appendix eq: KL divergence}, we have
\begin{align}
     u(\pmb{s},\pmb{a};\hat{{\pmb{\beta}}}_n,{\pmb{\beta}}^\star, \pi)&\leq \sqrt{2} \left(1+\frac{r_{max}}{1-\gamma}\right)^{1/2}\sqrt{D_{KL}\left(\mu({\cdot}|\pmb{s},\pmb{a}; {\pmb{\beta}}^\star)\Vert \mu({\cdot}|\pmb{s},\pmb{a}; \hat{\pmb{\beta}}_n)\right)} \nonumber\\
     &= \sqrt{2} \left(1+\frac{r_{max}}{1-\gamma}\right)^{1/2}\sqrt{
     \E\left[\mathcal{L}({\pmb{\beta}}^\star) - \ell(\pmb{s},\pmb{a},\pmb{s}^\prime;\hat{\pmb{\beta}}_n)\right]} \nonumber\\
     &=\mathcal{O}\left(\sqrt{\psi(t)}\right) \label{apendix eq: calibration convergence rate assumption}
\end{align}
where step~\eqref{apendix eq: calibration convergence rate assumption} follows the assumed rate of calibration of loss in (\ref{eq.calibrationLoss}). 
Because $\mathcal{S}$ and $\mathcal{A}$ are both closed and compact space, we have  $$\epsilon_u(\pi) =
\E_{(\pmb{s},\pmb{a})\sim d_{\mathcal{M}^d}^{\pi}(\pmb{s})\pi(\pmb{a}|\pmb{s})}[u(\pmb{s},\pmb{a}; \hat{{\pmb{\beta}}}_n, \pi)]
=\mathcal{O}\left(\sqrt{\psi(t)}\right).$$
We conclude that the rate of convergence is given by applying \eqref{eq: theo 4 conv}
\begin{equation*}
 { J(\pi^\star; \mathcal{M}^p)- J(\hat{\pi}_n; \mathcal{M}^p) \leq (\lambda+\gamma)\epsilon_u(\pi^\star)} = \mathcal{O}\left(\sqrt{\psi(t)}\right).
\end{equation*}
\hfill 
\endproof

\section{Estimation of Information Matrix}
\label{appendix: information matrix estimation}

Suppose the state transition model follows the general form of $\pmb{s}^\prime=f(\pmb{s},\pmb{a};\pmb\beta)+\pmb{\epsilon}$, where $f(\cdot,\cdot;\pmb\beta)$ is the deterministic mean function, 
$\pmb{s}$ is the current state, $\pmb{s}^\prime$ is the next state, and $\pmb{a}$ is the action. 
{The term $\pmb{\epsilon}$
follows a multivariate Gaussian distribution with 
zero mean and covariance matrix $\Sigma$. Thus the log-likelihood becomes $\log {\mu}(\pmb{s}^\prime|\pmb{s},\pmb{a};\pmb\beta)= -\frac{1}{2}(\pmb{s}^\prime-f(\pmb{s},\pmb{a};\pmb\beta))^{\top}\Sigma^{-1}(\pmb{s}^\prime-f(\pmb{s},\pmb{a};\pmb\beta))-\log(2\pi\Sigma)$. 
} 
We consider the conditional information matrix, 
$$
\mathcal{I}({\pmb\beta}|\pmb{s},\pmb{a})=-\E\left[\nabla^2 \log {\mu}(\pmb{s}^\prime|\pmb{s},\pmb{a};{\pmb{\beta}})\right].
$$ 
By taking second-order derivative of the log-likelihood with respect to $\pmb\beta$, we have 
\begin{eqnarray}
 -\E\left[\frac{\partial^2 \log {\mu}(\pmb{s}^\prime|\pmb{s},\pmb{a};\pmb\beta)}{\partial \pmb{\beta}\partial\pmb{\beta}} \right]
= \E\left[\left(\frac{\partial f(\pmb{s},\pmb{a};\pmb\beta)}{\partial \pmb\beta}\right)^{\top}\Sigma^{-1}\frac{\partial f(\pmb{s},\pmb{a};\pmb\beta)}{\partial \pmb\beta}- (\pmb{s}^\prime-f(\pmb{s},\pmb{a};\pmb\beta))\Sigma^{-1}\frac{\partial^2 f(\pmb{s},\pmb{a};\pmb\beta)}{\partial \pmb{\beta}\partial \pmb{\beta}}\right].
 \nonumber 
\end{eqnarray}
    Given the expectation that $\E[\pmb{s}^\prime-f(\pmb{s},\pmb{a};\pmb\beta)]=0$, the second term involving $\pmb{s}^\prime-f(\pmb{s},\pmb{a};\pmb\beta)$ averages to zero. By simplifying the expression, the conditional information matrix $[\mathcal{I}(\pmb\beta|\pmb{s},\pmb{a})]=\left(\frac{\partial f(\pmb{s},\pmb{a};\pmb\beta)}{\partial \pmb\beta}\right)^{\top}\Sigma^{-1}\frac{\partial f(\pmb{s},\pmb{a};\pmb\beta)}{\partial \pmb\beta}$.
The gradient of
the model output with respect to parameters $\pmb\beta$ can be computed through the model $f(\cdot,\cdot;\pmb\beta)$ using the automatic differentiation library. 

For the general exponential family, the information matrix can be estimated by using Lemma 3.1 of \cite{overstall2022properties}.
The Fisher information matrix considers the structural information in 
the dynamic model, then elucidates how uncertainties in parameter estimation propagate, ultimately influencing predictions and inferences.

\section{Experiment Setup}\label{appendix: experiment setup}

We conducted the empirical experiments using a metabolic reaction network simulator designed to replicate the dynamic behavior of cell cultures under various enzymatic and 
environmental conditions. 
The initial metabolic state was selected based on 
the iPSC culture experimental data derived from \cite{odenwelder2021induced}, Additionally, a 20\% perturbation was added to sample the initial state. 

\textbf{Simulation Configuration.} The metabolic network was configured to test various max enzymatic reaction rates (e.g., $v_{max,PDH}$, $v_{max,CS}$) and substrate affinities (e.g., $K_{m,GLC}$, $K_{m,GLU}$); see the detailed model in Appendix~\ref{appendix: mechanistic model} and \cite{wang2023metabolic}. The number of calibration parameters varied in three case study settings with $20$, $30$, and $40$ 
parameters.
{For each case study, we initialized the digital twin by running five experiments with random medium exchange ratios and each experiment trajectory comprising 12 timesteps. We then conducted 100 sequential physical experiments to collect additional data. 
{At every $n$-th iteration,  the historical dataset $\mathcal{D}_{n}$ is used to update the model parameter estimate $\hat{\pmb{\beta}}_n$. The conditional Fisher Information matrix ${\mathcal{I}(\hat{\pmb{\beta}}_{n};\pmb{s},\pmb{a})}$ and Hessian matrix $\Sigma(\hat{\pmb{\beta}}_{n})$ are further computed for estimating the uncertainty function \eqref{eq: weighted distance estimation} that is used in the calibration policy}. During preliminary exploration, we observed the training was vulnerable to overfitting due to high sample dependence. To mitigate this, we employed an early-stopping strategy. We performed 30 macro-replications to ensure robust performance metrics. Finally, we evaluated the test error (e.g., the Mean Absolute Percentage Error (MAPE)) by running the 
the estimated optimal policy on the physical system for 2,000 episodes.
The reward function parameters are set $c_r=30$ \$/g, $c_m=120$ \$/100\%, $c_l=84$ \$/g.}

{\textbf{Policy Optimization and Agent Configuration.} Given state $\pmb{s}$, to learn an optimal action $\pmb{a}$, a Deep Q-Network (DQN) \citep{mnih2015human} agent consists of a feed-forward neural network with two fully connected hidden layers, each containing 64 neurons, followed by ReLU activation functions. The input layer dimension corresponds to the size of the state space \(\texttt{input\_dims}\), and the output layer produces a Q-value, denoted by $Q(\pmb{s},\pmb{a}; \pmb{\theta})$, for each possible action (\(\texttt{n\_actions}\)). 
That means for a given state \(\pmb{s}\), the network computes $Q(\pmb{s},\pmb{a}; \pmb{\theta})$ for each possible $\pmb{a}$ in $\mathcal{A}$.}

The agent uses the Adam optimizer \citep{kingma2015adam} with a learning rate \(0.001\) and a small L2 regularization constant of \(10^{-10}\). 
The loss function is the Mean Squared Error (MSE) between the predicted Q-values and the TD target. To stabilize training, the agent maintains a replay buffer with a fixed maximum size (e.g., \(100{,}000\) transitions). At each timestep, the agent stores \((\pmb{s}, \pmb{a}, r, \pmb{s}')\) in the memory. For training, a mini-batch of size \(\texttt{batch\_size} = 64\) is sampled uniformly at random. The discount factor is set as $\gamma=0.99$.

\textbf{Epsilon-Greedy Exploration.} To balance exploration and exploitation, the agent employs an \(\epsilon\)-greedy policy. Formally, the agent’s action selection is given by:
\[
\pmb{a} = 
\begin{cases}
   \arg\max_{a'} Q(\pmb{s},\pmb{a}'; \pmb{\theta}) & \text{with probability } (1-\epsilon),\\
 \mbox{Uniform Sampling}
 & \text{with probability } \epsilon.
\end{cases}
\]
With probability \(\epsilon\), we select an action {$\pmb{a}=\pi(\pmb{s}; b)$ with $b\sim \mbox{Uniform} (0,1)$;} otherwise, the action with the highest predicted Q-value is chosen. Initially, \(\epsilon\) is set to 0.6, decaying linearly by \(\Delta \epsilon = 5\times 10^{-4}\) after each training step until it reaches a minimum, i.e., \(\epsilon_{\min} = 0.01\). This gradual decay allows sufficient exploration in the early stages of learning while converging to a more exploitative policy once the environment has been well-sampled. 


The agent was trained using a batch size of 64, iterating through the environmental states and adjusting actions based on  
{the corresponding rewards $r(\pmb{s},\pmb{a})$, and the uncertainty function, i.e., $u\left(\pmb{s},\pmb{a};\pmb{\beta},\pi\right)$}. 
Model training was guided by {maximum likelihood estimation (MLE). 
Action selection is directed by maximizing information gain, specifically selecting the action with the highest $u(\pmb{s},\pmb{a};\pmb{\beta},\pi)$.}
In addition, each experiment was seeded to ensure reproducibility, with results collected for actions, states, rewards, and optimized parameters over the course of the simulations. These results were stored systematically for further analysis and model parameter calibration.

\section{Mechanistic Model for iPSC Culture}\label{appendix: mechanistic model}

{The iPSC metabolic network and the Michaelis–Menten (M-M) formalized flux rate model are adapted from \cite{wang2023metabolic}. The network includes glycolysis, the TCA cycle, anaplerosis, amino acid metabolism, and a simplified PPP. The metabolic flux rates capture the effects of substrate and inhibitor concentrations. Additionally, regulatory mechanisms such as allosteric regulation, competitive inhibition, and feedback inhibition were incorporated into the metabolic flux kinetic model, as detailed here.}

\begin{table}[hbt!]
\centering
\caption{Biokinetic equations of the metabolites fluxes (1-30) of the model}
\label{tab:kinetic}
\begin{tabular}{|lp{12cm}|}
\hline
\textbf{No.} & \multicolumn{1}{c|}{\textbf{Pathway}}                    \\ \hline \hline
             & \multicolumn{1}{c|}{\textbf{Glycolysis}}   \\ \hline
\multicolumn{1}{|l|}{\textbf{1}}   & $v(HK)= {v_{max, HK}} \times \frac{Glc}{K_{m, Glc}+Glc} \times \frac{K_{i, G6P}}{K_{i, G6P}+G6P} \times \frac{K_{i, LactoHK}}{K_{i, LactoHK}+Lac}$    \\ \hline
\multicolumn{1}{|l|}{\textbf{2}}   & $v(PGI)= {v_{max, PGI}} \times \frac{G6P}{K_{m, G6P}+G6P}$   \\ \hline
\multicolumn{1}{|l|}{\textbf{3}}   & $v(PFK/ALD) = {v_{max, PFK/ALD}} \times \frac{F6P}{K_{m, F6P}+F6P}$                \\ \hline
\multicolumn{1}{|l|}{\textbf{4}}   & $v(PGK) = {v_{max, PGK}} \times \frac{GAP}{K_{m, GAP}+GAP}$   \\ \hline
\multicolumn{1}{|l|}{\textbf{5}}   & $v(PK) = {v_{max, PK}} \times \frac{PEP}{K_{m, PEP} \times (1+\frac{K_{a, F6P}}{F6P})+PEP}$          \\ \hline
\multicolumn{1}{|l|}{\textbf{6f}}    & $v{LDHf}= {v_{max, fLDH}} \times \frac{Pyr}{K_{m, Pyr}+Pyr}$                       \\ \hline
\multicolumn{1}{|l|}{\textbf{6r}}   & $v(LDHr) = {v_{max, rLDH}} \times \frac{Lac}{K_{m, Lac}+Lac} \times \frac{K_{i, Pyr}}{K_{i, Pyr}+Pyr}$   \\ \hline
\multicolumn{1}{|l|}{\textbf{7}}   & $v(PyrT) = {v_{max, PyrT}} \times \frac{EPyr}{K_{m, EPyr}+EPyr} \times \frac{K_{i, LactoPyr}}{K_{i, LactoPyr}+Lac}$              \\ \hline
\multicolumn{1}{|l|}{\textbf{8f}}   & $v(LacTf) = {v_{max, fLacT}} \times \frac{Lac}{K_{m, Lac}+Lac}$                     \\ \hline
\multicolumn{1}{|l|}{\textbf{8r}}    & $v(LacTr) = {v_{max, rLacT}} \times \frac{ELac}{K_{m, ELac}+ELac}$                  \\ \hline  \hline
\multicolumn{1}{|l|}{\textbf{}}    & \multicolumn{1}{c|}{\textbf{PPP}}                   \\ \hline
\multicolumn{1}{|l|}{\textbf{9}}   & $v(OP) = {v_{max, OP}} \times \frac{G6P}{K_{m, G6P}+G6P}$   \\ \hline
\multicolumn{1}{|l|}{\textbf{10}}   & $v(NOP) = {v_{max, NOP}} \times \frac{Ru5P}{K_{m, Ru5P}+Ru5P}$                     \\ \hline  \hline
\multicolumn{1}{|l|}{\textbf{}}    & \multicolumn{1}{c|}{\textbf{TCA}}                   \\ \hline
\multicolumn{1}{|l|}{\textbf{11}}   & $v(PDH) ={ v_{max, PDH}} \times \frac{Pyr}{K_{m, Pyr}+Pyr}$   \\ \hline
\multicolumn{1}{|l|}{\textbf{12}}    & $v(CS) ={ v_{max, CS}} \times \frac{AcCoA}{K_{m, AcCoA}+AcCoA} \times \frac{OAA}{K_{m, OAA}+OAA}$      \\ \hline
\multicolumn{1}{|l|}{\textbf{13f}}   & $v(CITS/ISODf) = {v_{max, fCITS/ISOD}} \times \frac{Cit}{K_{m, Cit}+Cit}$          \\ \hline
\multicolumn{1}{|l|}{\textbf{13r}}   & $v(CITS/ISODr) = {v_{max, rCITS/ISOD}} \times \frac{AKG}{K_{m, AKG}+AKG}$          \\ \hline
\multicolumn{1}{|l|}{\textbf{14}}   & $v(AKGDH) = {v_{max, AKGDH}} \times \frac{AKG}{K_{m, AKG}+AKG}$                     \\ \hline
\multicolumn{1}{|l|}{\textbf{15}}    & $v(SDH) = {v_{max, SDH}} \times \frac{Suc}{K_{m, SUC}+Suc}$                        \\ \hline
\multicolumn{1}{|l|}{\textbf{16f}}                        & $v(FUMf) = {v_{max, fFUM}} \times \frac{Fum}{K_{m, Fum}+Fum}$                      \\ \hline
\multicolumn{1}{|l|}{\textbf{16r}}   & $v(FUMr) = {v_{max, rFUM}} \times \frac{Mal}{K_{m, Mal}+Mal}$                     \\ \hline
\multicolumn{1}{|l|}{\textbf{17f}}                        & $v(MDHf) = {v_{max, fMDH}} \times \frac{Mal}{K_{m, Mal}+Mal}$                       \\ \hline
\multicolumn{1}{|l|}{\textbf{17r}}   & $v(MDHr) = {v_{max, rMDH}} \times \frac{OAA}{K_{m, OAA}+OAA}$                       \\ \hline
\end{tabular}
\end{table}

\begin{table}[t!]
\centering
\begin{tabular}{|lp{12cm}|}
\hline
\multicolumn{1}{|l|}{\textbf{}}    & \multicolumn{1}{c|}{\textbf{Anaplerosis and Amino Acid}}                   \\ \hline
\multicolumn{1}{|l|}{\textbf{18}}    & $v(ME) = {v_{max, ME}} \times \frac{Mal}{K_{m, Mal}+Mal}$                        \\ \hline
\multicolumn{1}{|l|}{\textbf{19}}   & $v(PC) = {v_{max, PC}} \times \frac{Pyr}{K_{m, Pyr}+Pyr}$                        \\ \hline
\multicolumn{1}{|l|}{\textbf{20f}}   & $v(GLNSf) = {v_{max, fGLNS}} \times \frac{Gln}{K_{m, Gln}+Gln} \times \frac{K_{i, LactoGLNS}}{K_{i, LactoGLNS}+Lac}$             \\ \hline
\multicolumn{1}{|l|}{\textbf{20r}}                        & $v(GLNSr) = {v_{max, rGLNS}} \times \frac{Glu}{K_{m, Glu}+Glu} \times \frac{NH_4}{K_{m, NH_4}+NH_4}$      \\ \hline
\multicolumn{1}{|l|}{\textbf{21f}}   & $v(GLDHf) = {v_{max, fGLDH}} \times \frac{Glu}{K_{m, Glu}+Glu}$                     \\ \hline
\multicolumn{1}{|l|}{\textbf{21r}}                        & $v(GLDHr) = {v_{max, rGLDH}} \times \frac{AKG}{K_{m, AKG}+AKG} \times \frac{NH_4}{K_{m, NH_4}+NH_4}$      \\ \hline
\multicolumn{1}{|l|}{\textbf{22f}}   & $v(AlaTAf) = {v_{max, fAlaTA}} \times \frac{Glu}{K_{m, GLU}+Glu} \times \frac{Pyr}{K_{m, Pyr}+Pyr}$    \\ \hline
\multicolumn{1}{|l|}{\textbf{22r}}                        & $v(AlaTAr) = {v_{max, rAlaTA}} \times \frac{Ala}{K_{m, Ala}+Ala} \times \frac{AKG}{K_{m, AKG}+AKG} \times (1+\frac{K_{a, Gln}}{Gln})$             \\ \hline
\multicolumn{1}{|l|}{\textbf{23}}    & $v(AlaT) = {v_{max, AlaT}} \times \frac{Ala}{K_{m, Ala}+Ala}$                       \\ \hline
\multicolumn{1}{|l|}{\textbf{24}}   & $v(GluT) = {v_{max, GluT}} \times \frac{Glu}{K_{m, Glu}+Glu}$                       \\ \hline
\multicolumn{1}{|l|}{\textbf{25}}    & $v(GlnT) = {v_{max, GlnT}} \times \frac{EGln}{K_{m, EGln}+EGln} \times \frac{K_{i, GLN}}{K_{i, GLN}+GLN}$                        \\ \hline
\multicolumn{1}{|l|}{\textbf{26}}   & $v(SAL) = {v_{max, SAL}} \times \frac{Ser}{K_{m, Ser}+Ser}$                \\ \hline
\multicolumn{1}{|l|}{\textbf{27f}}   & $v(ASTAf) = {v_{max, fASTA}} \times \frac{Asp}{K_{m, ASP}+Asp} \times \frac{AKG}{K_{m, AKG}+AKG}$         \\ \hline
\multicolumn{1}{|l|}{\textbf{27r}}   & $v(ASTAr) = {v_{max, rASTA}} \times \frac{Glu}{K_{m, Glu}+Glu} \times \frac{OAA}{K_{m, OAA}+OAA} \times \frac{NH_4}{K_{m, NH_4}+NH_4}$                 \\ \hline
\multicolumn{1}{|l|}{\textbf{28}}    & $v(AspT) = {v_{max, AspT}} \times \frac{EAsp}{K_{m, EAsp}+EAsp}$                       \\ \hline
\multicolumn{1}{|l|}{\textbf{29}}  & $v(ACL) = {v_{max, ACL}} \times \frac{Cit}{K_{m, Cit}+Cit}$    \\ \hline  \hline
\multicolumn{1}{|l|}{}             & \multicolumn{1}{c|}{\textbf{Biomass}}               \\ \hline
\multicolumn{1}{|l|}{\textbf{30}}  & $v(growth) = {v_{max, growth}} \times \frac{Gln}{K_{m, Gln}+Gln} \times \frac{Glc}{K_{m, Glc}+Glc} \times \frac{Glu}{K_{m, Glu}+Glu} \times \frac{Ala}{K_{m, Ala}+Ala} \times \frac{Asp}{K_{m, Asp}+Asp} \times \frac{Ser}{K_{m, Ser}+Ser} \times \frac{Gly}{K_{m, Gly}+Gly}$ \\ \hline
\end{tabular}
\end{table}

\end{appendices}

\clearpage

\bibliography{sn}

\end{document}